\documentclass[nohyperref]{article}

\usepackage{microtype}
\usepackage{graphicx}
\usepackage{subcaption}
\usepackage{booktabs}
\usepackage{multirow}

\usepackage{hyperref}

\usepackage[accepted]{icml2022}

\usepackage{amsmath}
\usepackage{amssymb}
\usepackage{mathtools}
\usepackage{amsthm}
\usepackage{amsmath} 
\usepackage{tikz}
\usepackage{etoolbox} 
\usepackage{listofitems} 
\usetikzlibrary{arrows.meta} 
\usetikzlibrary{patterns}
\usepackage[outline]{contour} 
\contourlength{1.4pt}

\tikzset{>=latex} 
\usepackage{xcolor}
\colorlet{myred}{red!80!black}
\colorlet{myblue}{blue!80!black}
\colorlet{myyellow}{yellow!80!black}
\colorlet{mygreen}{green!60!black}
\colorlet{myorange}{orange!70!red!60!black}
\colorlet{mydarkred}{red!30!black}
\colorlet{mydarkblue}{blue!40!black}
\colorlet{mydarkgreen}{green!30!black}
\tikzstyle{node}=[thick,circle,draw=myblue,minimum size=22,inner sep=0.5,outer sep=0.6]
\tikzstyle{node in}=[node,green!15!black,draw=mygreen,fill=mygreen!25]
\tikzstyle{node hidden}=[node,yellow!15!black,draw=myyellow,fill=myyellow!20]
\tikzstyle{node hidden2}=[node,blue!15!black,draw=myblue,fill=myblue!20]
\tikzstyle{node convol}=[node,orange!15!black,draw=brown,fill=brown!20]
\tikzstyle{node out}=[node,red!15!black,draw=myred,fill=myred!20]
\tikzstyle{connect}=[thick,mydarkblue] 
\tikzstyle{connect arrow}=[-{Latex[length=4,width=3.5]},thick,mydarkblue,shorten <=0.5,shorten >=1]
\tikzset{ 
  node 0/.style={node convol},
  node 1/.style={node in},
  node 2/.style={node hidden},
  node 3/.style={node out},
}

\pgfdeclarepatternformonly{soft north east lines}{\pgfqpoint{-1pt}{-1pt}}{\pgfqpoint{4pt}{4pt}}{\pgfqpoint{3pt}{3pt}}%
{
    \pgfsetstrokeopacity{0.5}
  \pgfsetlinewidth{0.1pt}
  \pgfpathmoveto{\pgfqpoint{0pt}{0pt}}
  \pgfpathlineto{\pgfqpoint{3.1pt}{3.1pt}}
  \pgfusepath{stroke}
}

\usepackage[capitalize,noabbrev]{cleveref}

\theoremstyle{plain}

\newtheorem{property}{Property}[section]

\theoremstyle{definition}

\theoremstyle{remark}

\usepackage[textsize=tiny]{todonotes}

\icmltitlerunning{Transformer Neural Processes}

\begin{document}

\twocolumn[
\icmltitle{Transformer Neural Processes: \\ Uncertainty-Aware Meta Learning Via Sequence Modeling}

\icmlsetsymbol{equal}{*}

\begin{icmlauthorlist}
\icmlauthor{Tung Nguyen}{ucla}
\icmlauthor{Aditya Grover}{ucla}
\end{icmlauthorlist}

\icmlaffiliation{ucla}{Department of Computer Science, UCLA}

\icmlcorrespondingauthor{Tung Nguyen}{tungnd@cs.ucla.edu}

\icmlkeywords{Transformers, Uncertainty, Meta-learning, Sequence modeling, Decision Making, Machine Learning, ICML}

\vskip 0.3in
]

\printAffiliationsAndNotice{}

\begin{abstract}
Neural Processes (NPs) are a popular class of approaches for meta-learning. 
Similar to Gaussian Processes (GPs), NPs define distributions over functions 
and can estimate uncertainty in their predictions.
However, unlike GPs, NPs and their variants suffer from underfitting and often have intractable likelihoods, which limit their applications in sequential decision making. 
We propose Transformer Neural Processes (TNPs), a new member of the NP family that casts uncertainty-aware meta learning as a sequence modeling problem.
We learn TNPs via an autoregressive likelihood-based objective and instantiate it with a novel transformer-based architecture. 
The model architecture respects the inductive biases inherent to the problem structure, such as invariance to the observed data points and equivariance to the unobserved points.
We further investigate knobs within the TNP framework that tradeoff expressivity of the decoding distribution with extra computation.
Empirically, we show that TNPs achieve state-of-the-art performance on various benchmark problems, outperforming all previous NP variants on meta regression, image completion, contextual multi-armed bandits, and Bayesian optimization.

\end{abstract}

\section{Introduction}

The goal of meta learning \citep{schmidhuber1987evolutionary,vanschoren2018meta} is to learn models that can adapt quickly to unseen tasks with only a few labeled examples. Since the amount of labeled data for any new task is limited, we ideally require the model to have both high accuracy and quantify the uncertainty in its predictions.
This is particularly important within sequential decision making, e.g., for Bayesian optimization \citep{mockus1978application,schonlau1998global,shahriari2015taking,hakhamaneshi2021jumbo} and multi-armed bandits \citep{cesa2006prediction,riquelme2018deep}, where the quantified uncertainty can guide data acquisition. We call this paradigm uncertainty-aware meta learning, which is the main focus of our paper.

Neural Processes (NPs) \citep{garnelo2018conditional,garnelo2018neural} are a rich class of models for such problems. 
NPs offer a flexible way to modeling distributions over functions, and are trained via a meta-learning framework that enables rapid adaptation to new functions at test time. More specifically, NPs are structured similar to a Variational Autoencoder (VAE) \citep{kingma2013auto} over datasets, wherein we employ a latent variable model to estimate the conditional distribution over the labels of the unlabeled (context) points given the set of labeled (target) points.
By amortizing the functional uncertainty in the encoded latent variables, NPs can be quickly adapted to a new task at test-time.

However, NPs and their variants suffer from two major drawbacks.
First, many NP variants often have intractable marginal likelihoods due to the presence of latent variables and instead optimize for a surrogate variational lower bound of the log-likelihood. 
The standard justification for using latent variables is that they can represent functional uncertainty and improve the predictive performance in some cases~\cite{le2018empirical}.
However, it is also known that optimizing variational lower bounds of the log-likelihood does not necessarily lead to a meaningful latent representation and can require non-trivial adjustments to the objective~\citep{chen2016variational,alemi2018fixing}. 
Second, it has been observed that NPs tend to underfit to the data distribution.  Attentive Neural Processes (ANPs) \citep{kim2019attentive} partly address this problem by incorporating attention mechanisms into the NP encoder-decoder architecture. 
While ANPs provide a considerably improved fit, these models tend to make overconfident predictions and have poor performance on sequential decision making problems.

We propose Transformer Neural Processes (TNPs), a new framework for uncertainty-aware meta learning derived from a sequence modeling perspective.
The learning objective for TNPs is to autoregressively maximize the conditional log-likelihood of the target points (observed only during training) conditioned on the context points (observed during training and testing).
This removes the need for latent variables and any variational approximations, while allowing for an expressive parameterization of the predictive distribution.
We instantiate TNPs via a transformer-based architecture with a causal mask, similar to GPT-x models~\citep{radford2018improving,radford2019language,brown2020language}.
Such models have led to state-of-the-art performance of a wide variety of domains and modalities~\cite{brown2020language,lu2021pretrained,chen2021decision}.
However, vanilla transformers cannot be directly used to parameterize a Neural Process as they lack invariance to the conditioned tokens and are not equivariant to the ordering of the target points.
We propose suitable modifications to satisfy these desiderata by removing positional embeddings, using a novel padding and masking scheme for the context and target points, and considering a Monte Carlo approximation to a symmetrized predictive distribution that is equivariant by design.

Finally, we also propose two variants of TNPs, which can tradeoff the expressivity of the autoregressive factorization with computational tractability and exact equivariance. These variants consider diagonal and Cholesky approximations to the covariance matrix of the output distribution.
Empirically, we evaluate TNPs on various benchmark problems proposed in prior works, including meta regression, image completion, contextual bandits, and Bayesian Optimization (BO). While meta regression and image completion evaluate the quality of predictions produced by TNPs, contextual bandits and BO directly measure the performance of TNPs on sequential decision making tasks. 
On all these problems, we observe that TNPs outperform several attention~\cite{kim2019attentive,lee2020bootstrapping} and non-attention based variants~\cite{garnelo2018conditional,garnelo2018neural} of NPs by a large margin.

\section{Background}
\subsection{Uncertainty-Aware Meta Learning}

In meta learning, we assume an unknown distribution over functions, say $\mathcal{F}$.
During training, we sample a fixed number of functions from $\mathcal{F}$ and observe a finite set of evaluations $\{x_i, y_i\}_{i=1}^N$ from each function $f: \mathcal{X} \rightarrow \mathcal{Y}$. 
At test time, we evaluate the generalization ability of the model on a set of unseen functions, assumed to be drawn from the same or similar distribution as $\mathcal{F}$. For each test function, we provide a small set of labelled training points $D_\text{train}$, and test the ability of the model to make predictions on a test set $D_\text{test}$.

Our focus in this work is on uncertainty-aware meta learning. That is, we assume that the model outputs a joint predictive distribution over the entire test set $D_\text{test}$.
For example, if we assume the predictive distribution is a multi-variate Gaussian with a diagonal covariance, we output a mean $\mu_j$ and standard deviation $\sigma_j$ for each input $x_j \in D_\text{test}$. Here, the $\sigma_j$'s quantify the uncertainty in the model's predictions.

\subsection{Neural Processes}
A Neural Process (NP) is a stochastic process that describes the predictive distribution over a set of unlabelled points (target) given a training set of labelled points (context)~\cite{garnelo2018conditional,garnelo2018neural}. 
Additionally, NPs incorporate a latent variable $z$ to represent the functional uncertainty.
Formally, the likelihood of the NP model is given as:
\begin{equation}
\begin{aligned}
    & p(y_{m+1:N} \vert x_{m+1:N}, C) \\
    = & \int_{z} p(y_{m+1:N} \vert x_{m+1:N}, z) p(z \vert C) dz,
\end{aligned}
\end{equation}
in which $C = \{x_i,y_i\}_{i=1}^m$ and $T = \{x_i, y_i\}_{i=m+1}^N$ is the set of context and target pairs respectively. 
We note that there also exists variants of neural processes with no latent variables; we refer the reader to Section~\ref{sec:related} for a detailed review.
As the likelihood is intractable, NPs maximize an evidence lower bound (ELBO) of the log-likelihood instead:
\begin{equation}
\resizebox{0.99\hsize}{!}{%
$
\begin{aligned}
 &\log p(y_{m+1:N} \vert x_{m+1:N}, C) \geq  \\
    &\mathbb{E}_{q(z \vert C, T)}[\log p(y_{m+1:N} \vert x_{m+1:N}, z)]   - \mathrm{KL}(q(z \vert C, T) || p(z \vert C)).
\end{aligned}
$
}
\end{equation}
Equivalently, we can view NPs as a VAE \citep{kingma2013auto} over the target labels conditioned on the context pairs and the target inputs.
The encoder $q(z \vert C, T)$ is a permutation-invariant function which maps the context pairs to a distribution over $z$. In practice, $q$ consists of an MLP that maps each pair $(x_i,y_i)$ to its representation, an aggregator that combines these representations, and another MLP that outputs the mean and variance of $z$. The decoder $p(y_{m+1:N} \vert x_{m+1:N}, z) = \prod_{i=m+1}^N p(y_i \vert x_i, z)$ predicts the label for each target independently given the inferred $z$.

\subsection{Transformers}
Transformers were proposed by \citet{vaswani2017attention} as an efficient architecture for modeling sequential data. Transformers are composed of an encoder and an decoder, which both consist of a stack of self-attention layers with residual connections. The self-attention layer receives $n$ embeddings $\{e^{\text{in}}_i\}_{i=1}^n$, which are associated with $n$ input tokens, and outputs $n$ corresponding embeddings $\{e^{\text{out}}_i\}_{i=1}^n$. Each embedding vector $e^{\text{in}}_i$ is first mapped to a key $k_i$, query $q_i$, and value $v_i$ via linear transformations. The output embedding $e^{\text{out}}_i$ is a weighted linear combination of all the input values, where the weights are given by the normalized dot-product between its query $q_i$ and other keys $k_j$:
\begin{equation}
    e^{\text{out}}_i = \sum_{j=1}^n \frac{\exp (\langle q_i, k_j \rangle)}{\sum_{j'=1}^n \exp (\langle q_i, k_{j'} \rangle)} \cdot v_j.
\end{equation}
Before the first self-attention layer, each input token is passed through a positional encoder, which incorporates sequential information into the input sequence. This flexible architecture allows transformers to handle input sequences of arbitrary lengths, and provides a simple yet efficient way to model the relationships between input tokens. This architecture has been the key to many recent breakthroughs in language and vision~\citep{radford2019language,dosovitskiy2020image,chen2020generative,brown2020language}. In this paper, we study how transformers can be applied to uncertainty-aware meta-learning problems.

\section{Transformer Neural Processes} \label{sec:sequence}

We propose to solve uncertainty-aware meta learning via the lens of sequence modeling. In order to do so, we consider each set of evaluations $\{x_i,y_i\}_{i=1}^N$ that the model observes during training as an ordered sequence of $N$ data points.
Since we observe evaluations from multiple such functions, we segregate a random subset of each training sequence as the set of context pairs $(x_{1:m}, y_{1:m})$ as few-shot conditioning for the sequence model.
Thereafter, we autoregressively model the predictive likelihood of the remaining $(N-m)$ target points and maximize the following objective:
\begin{align}
    \mathcal{L}&( \theta) = \mathbb{E}_{x_{1:N}, y_{1:N}, m}\left[\log p_\theta(y_{m+1:N} \mid x_{1:N}, y_{1:m})\right] \label{eq:objective} \\
    & = \mathbb{E}_{x_{1:N}, y_{1:N}, m} \left[\sum_{i=m+1}^N \log p_\theta(y_i \mid x_{1:i}, y_{1:i-1}) \right] \label{eq:chain_rule}.
\end{align}
Each conditional in the above objective is a univariate Gaussian distribution.
In practice, we optimize a Monte Carlo approximation of the objective in Eq.~\ref{eq:chain_rule}, where we consider a batch of training functions and their randomly sampled evaluations. We uniformly sample an index $m$ that determines the context and target points for each set of evaluations.
Next, we list the desiderata for the architectures that will be used for parameterizing the sequence model. 

\begin{property}
\textbf{Context invariance.} A model $p_\theta$ is context invariant if for any choice of permutation function $\pi$ and $m\in [1, N-1]$, $p_\theta(y_{m+1:N} \mid x_{m+1:N}, x_{1:m}, y_{1:m})$ $=$ $p_\theta(y_{m+1:N} \mid x_{m+1:N}, x_{\pi(1):\pi(m)}, y_{\pi(1):\pi(m)})$. \label{property_1}
\end{property}
\begin{property}
\textbf{Target equivariance.} A model $p_\theta$ is target equivariant if for any choice of permutation function $\pi$ and $m\in [1, N-1]$, $p_\theta(y_{m+1:N} \mid x_{m+1:N}, x_{1:m}, y_{1:m})$ $=$ $p_\theta(y_{\pi(m+1):\pi(N)} \mid x_{\pi(m+1):\pi(N)}, x_{1:m}, y_{1:m})$. \label{property_2}
\end{property}

\begin{figure*}
\centering
\begin{tikzpicture}[scale=0.7, every node/.style={scale=0.8}]

\def\xa{1.5}
\def\ya{0}
\def\xb{\xa+4}
\def\xc{\xb+7.5}
\def\xd{\xc+7.5}
\def\yb{\ya+3}
\def\yc{\yb+1.5}

\node[node 1, minimum size=28] (x_1) at (\xa+0, \ya) {$x_1$};
\node[rectangle, draw, rounded corners, minimum width=2.5cm,minimum height=0.75cm, fill=myyellow!50, align=center] (embx_1) at (\xa+0.75, \ya+1.6){Embed};
\node[node 3, minimum size=28] (y_1) at (\xa+1.5, \ya) {$y_1$};
\node[rectangle, draw, dashed, rounded corners, minimum width=2.75cm,minimum height=1.25cm] (box_1) at (\xa+0.75, \ya){};

\filldraw[black] (\xa+2.5,\ya) circle (1pt);
\filldraw[black] (\xa+2.75,\ya) circle (1pt);
\filldraw[black] (\xa+3,\ya) circle (1pt);

\node[node 1, minimum size=28] (x_m) at (\xb+0, \ya) {$x_m$};
\node[rectangle, draw, rounded corners, minimum width=2.5cm,minimum height=0.75cm, fill=myyellow!50, align=center] (embx_m) at (\xb+0.75, \ya+1.6){Embed};
\node[node 3, minimum size=28] (y_m) at (\xb+1.5, \ya) {$y_m$};
\node[rectangle, draw, dashed, rounded corners, minimum width=2.75cm,minimum height=1.25cm] (box_m) at (\xb+0.75, \ya){};
\node[] (context) at (\xa+2.75,\ya-1){Context points};

\node[node 1, minimum size=28] (x_m+1) at (\xb+3.5, \ya) {$x_{m+1}$};
\node[rectangle, draw, rounded corners, minimum width=2.5cm,minimum height=0.75cm, fill=myyellow!50, align=center] (embx_m+1) at (\xb+4.25, \ya+1.6){Embed};
\node[node 3, minimum size=28] (y_m+1) at (\xb+5, \ya) {$y_{m+1}$};
\node[rectangle, draw, dashed, rounded corners, minimum width=2.75cm,minimum height=1.25cm] (box_m+1) at (\xb+4.25, \ya){};

\filldraw[black] (\xb+6,\ya) circle (1pt);
\filldraw[black] (\xb+6.25,\ya) circle (1pt);
\filldraw[black] (\xb+6.5,\ya) circle (1pt);

\node[node 1, minimum size=28] (x_N) at (\xc+0, \ya) {$x_N$};
\node[rectangle, draw, rounded corners, minimum width=2.5cm,minimum height=0.75cm, fill=myyellow!50, align=center] (embx_N) at (\xc+0.75, \ya+1.6){Embed};
\node[node 3, minimum size=28] (y_N) at (\xc+1.5, \ya) {$y_N$};
\node[rectangle, draw, dashed, rounded corners, minimum width=2.75cm,minimum height=1.25cm] (box_N) at (\xc+0.75, \ya){};
\node[black] (target) at (\xb+6.25,\ya-1){Target points};

\node[node 1, minimum size=28] (x_m+1_2) at (\xc+3.5, \ya) {$x_{m+1}$};
\node[rectangle, draw, rounded corners, minimum width=2.5cm,minimum height=0.75cm, fill=myyellow!50, align=center] (embx_m+1_2) at (\xc+4.25, \ya+1.6){Embed};
\node[node 3, minimum size=28] (y_m+1_2) at (\xc+5, \ya) {$0$};
\node[rectangle, draw, dashed, rounded corners, minimum width=2.75cm,minimum height=1.25cm] (box_m+1_2) at (\xc+4.25, \ya){};

\filldraw[black] (\xc+6,\ya) circle (1pt);
\filldraw[black] (\xc+6.25,\ya) circle (1pt);
\filldraw[black] (\xc+6.5,\ya) circle (1pt);

\node[node 1, minimum size=28] (x_N) at (\xd+0, \ya) {$x_N$};
\node[rectangle, draw, rounded corners, minimum width=2.5cm,minimum height=0.75cm, fill=myyellow!50, align=center] (embx_N_2) at (\xd+0.75, \ya+1.6){Embed};
\node[node 3, minimum size=28] (y_N_2) at (\xd+1.5, \ya) {$0$};
\node[rectangle, draw, dashed, rounded corners, minimum width=2.75cm,minimum height=1.25cm] (box_N_2) at (\xd+0.75, \ya){};
\node[black] (padded) at (\xc+6.25,\ya-1){Padded target points};

\node[rectangle, draw,rounded corners, minimum width=20.5cm,minimum height=1.0cm, fill=myblue!20, align=center] (tr) at (\xb+6.25, \yb){Transformer Neural Processes};

\node[node 3, minimum size=28] (yhat_m+1) at (\xc+4.25, \yc) {$\hat{y}_{m+1}$};
\filldraw[black] (\xc+5.75,\yc) circle (1pt);
\filldraw[black] (\xc+6.25,\yc) circle (1pt);
\filldraw[black] (\xc+6.75,\yc) circle (1pt);
\node[node 3, minimum size=28] (yhat_N) at (\xd+0.75, \yc) {$\hat{y}_{N}$};

\draw [->] (box_1) to  (embx_1 |- embx_1.south);
\draw [->] (box_m) to  (embx_m |- embx_m.south);
\draw [->] (box_m+1) to  (embx_m+1 |- embx_m+1.south);
\draw [->] (box_N) to  (embx_N |- embx_N.south);
\draw [->] (box_m+1_2) to  (embx_m+1_2 |- embx_m+1_2.south);
\draw [->] (box_N_2) to  (embx_N_2 |- embx_N_2.south);

\draw [->] (embx_1) to  (embx_1 |- tr.south);
\draw [->] (embx_m) to  (embx_m |- tr.south);
\draw [->] (embx_m+1) to  (embx_m+1 |- tr.south);
\draw [->] (embx_N) to  (embx_N |- tr.south);
\draw [->] (embx_m+1_2) to  (embx_m+1_2 |- tr.south);
\draw [->] (embx_N_2) to  (embx_N_2 |- tr.south);

\draw [<-] (yhat_m+1) to (yhat_m+1 |- tr.north);
\draw [<-] (yhat_N) to (yhat_N |- tr.north);



\end{tikzpicture}

\caption{Illustration of the TNP-A architecture. The architecture specifies a custom masking pattern between the contexts, targets, and padded targets to respect autoregressive prediction order. For TNP-D and TNP-ND, we remove the targets from the input sequence.
}
\label{fig:TNP_general}
\vspace{-0.15in}
\end{figure*}
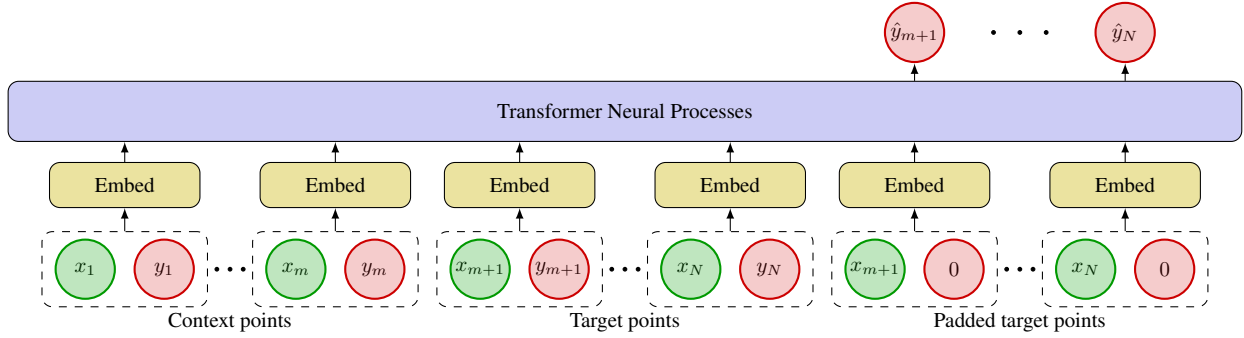

Context invariance (Property~\ref{property_1}) requires that for any underlying function $f$, the predictions for the target points should not change if we permute the context points.
Target equivariance (Property~\ref{property_2}) requires that whenever we permute the target inputs $\{x_i\}_{i=m+1}^N$, the predictions are permuted accordingly. 
We instantiate the objective in Eq.~\ref{eq:chain_rule} with a transformer architecture as shown in Figure \ref{fig:TNP_general}.
Standard transformers such as GPT~\cite{radford2018improving} employ a casual mask to enforce the autoregressive structure needed for Eq.~\ref{eq:chain_rule}.
However, this naive application of autoregressive transformers fails to satisfy the desiderata in Property~\ref{property_1} and Property~\ref{property_2}.
Since this model treats $x_i$ and $y_i$ as two separate tokens, we need to add them with a positional embedding vector for the model to associate them as a pair $(x_i, y_i)$. This positional encoding, unfortunately, makes the model's output depend on the permutation of the target points.
Further, autoregressive transformers also violate target equivariance as the ordering of points influences their embeddings and hence, different orderings lead to non-equivariant predictions.
Next, we present Transformer Neural Processes (TNPs), a family of transformer-based neural processes that mitigates the aforementioned challenges.

\subsection{Autoregressive Transformer Neural Process} \label{sec:tnp_a}
Our first attempt at TNPs builds on the autoregressive factorization in Eq.~\ref{eq:chain_rule}.
A vanilla transformer violates Property \ref{property_1} due to its use of positional encodings.
However, these encodings are also useful for drawing associations between $x_i$ and $y_i$.
To address this challenge, we first concatenate $x_i$ and $y_i$ to form a single token, 
which allows us to remove positional encodings and yet treat them as a pair.
While this scheme works well for the context points $(x_i, y_i)_{i=1}^m$, we cannot apply it naively for
any target point $y_{i>m}$ that depends on previous pairs $(x_j,y_j)_{j=1}^{i-1}$ and its input $x_i$.
To respect the autoregressive structure for such points, 
we introduce auxiliary tokens consisting of $x_{i>m}$ padded with a dummy token ($0$ in our case) and append them to our original sequence. 
Our padded sequence consists of $N$ real pairs $(x_i, y_i)_{i=1}^N$ and $N-m$ padded pairs $(x_i, 0)_{i=m+1}^N$:
\begin{align}\label{eq:tnp_seq}
 \resizebox{0.9\hsize}{!}{%
        $
    \tau = 
    \{
    (x_1, y_1), \dots, (x_{N}, y_{N}), (x_{m+1}, 0), \dots, (x_N, 0) 
    \}.
    $
}
\end{align}
To preserve the autoregressive ordering in (\ref{eq:chain_rule}), we design a masking mechanism in the attention layer such that:  \\
(1) the context points $(x_i, y_i)_{i=1}^m$ only attend to themselves; \\
(2) the target point  $(x_i, y_i)$ for $(i>m)$ attends to all context points and the previous target points $(x_j, y_j)_{j=m+1}^{i}$; \\
(3) the padded target point $(x_i,0)$ for $(i>m)$ attends to all context points and the previous target points $(x_j, y_j)_{j=m+1}^{i-1}$.  
\begin{figure}[h]
\centering
\includegraphics[width=0.75\columnwidth]{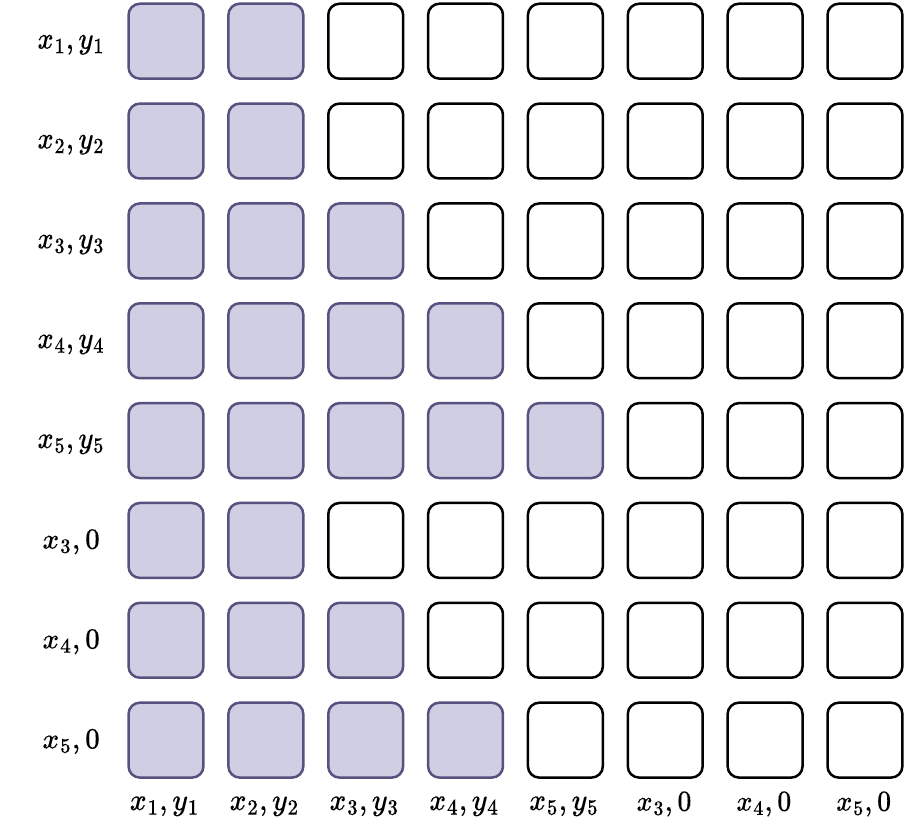}
\caption{An example mask with $N=5$ and $m=2$. Each token is allowed to attend to other filled tokens on its corresponding row. The context points $(x_i, y_i)_{i=1}^2$ attend to themselves. Each target point $(x_i, y_i)$ for $i > 2$ attends to the context points and the previous target points $(x_j,y_j)_{j=3}^i$. Each padded target point $(x_i, 0)$ for $i > 0$ attends to the context points and the previous target points $(x_j,y_j)_{j=3}^{i-1}$. This ensures the prediction for $y_i$ ($i>2$) only depends on the context and the previous target points. \vspace{-0.1cm}}
\label{fig:tnpa_mask}
\end{figure}
We refer to this model as Autoregressive Transformer Neural Processes (TNP-A). 
Figure~\ref{fig:TNP_general} illustrates the model, and Figure~\ref{fig:tnpa_mask} shows an example mask with $N=5$ and $m=2$.
It satisfies Property~\ref{property_1} as a consequence of the masking scheme described above.
For satisfying Property~\ref{property_2}, we follow a symmetrization argument. Here, we note that any function can be made equivariant to a group by averaging the function evaluations over the entire group~\cite{murphy2018janossy}.
Formally, let us denote our joint distribution of interest as  $\tilde{p}_\theta(y_{m+1:N} \mid x_{1:N}, y_{1:m})$.
We define $\tilde{p}_\theta$ in terms of a base autoregressive model $p_\theta$ as:
\begin{align}
    &\tilde{p}_\theta(y_{m+1:N} \vert x_{1:N}, y_{1:m}) \nonumber \\
    &= \mathbb{E}_\pi [p_\theta(y_{\pi(m+1):\pi(N)} \vert x_{\pi(m+1):\pi(N)}, x_{1:m}, y_{1:m})].
\end{align}
Since the permutation group is intractable to enumerate, we instead consider a Monte Carlo average over randomly sampled permutations to approximate $\tilde{p}_\theta$.
Even though the use of Monte Carlo implies that we satisfy Property~\ref{property_2} only in the limit, we can compute the predictive distribution tractably during training and evaluation.
Next, we introduce alternate decoders for TNPs that exactly satisfy Property~\ref{property_2} while trading off expressivity for computational tractability.

\subsection{Diagonal Transformer Neural Process}
We can simplify the objective in (\ref{eq:chain_rule}) by assuming that the target points $y_{m+1:N}$ are conditionally independent given the context points and the inputs $x_{m+1:N}$.
In other words, we consider the factorization:
\begin{equation}
\resizebox{0.95\hsize}{!}{%
    $
    p_\theta(y_{m+1:N} \vert x_{1:N}, y_{1:m}) = \prod_{i=m+1}^N  p_\theta(y_i \vert x_i, x_{1:m}, y_{1:m}).
    $
}
\end{equation}
As the target points are independent, we remove $\{(x_i, y_i)_{i=m+1}^N\}$ from the input sequence in (\ref{eq:tnp_seq}) and only feed the sequence consisting of the context points and the padded target points $\{(x_i, 0)_{i=m+1}^N\}$. 
This decoding distribution can be seen as a multivariate normal distribution with a diagonal covariance matrix and hence, we refer to it as Diagonal Transformer Neural Processes (TNP-D). 
Unlike TNP-A, we do not need to average over permutations to satisfy Property \ref{property_2}. 
However, for many scenarios, the independence assumption between the target points can be very strong for accurately modeling the underlying function. 

\subsection{Non-Diagonal Transformer Neural Process} \label{sec:tnp_nd}
Finally, we introduce Non-Diagonal Transformer Neural Processes (TNP-ND), the third variant of TNPs that balances between the tractability of TNP-D and the expressivity of TNP-A while satisfying both Property~\ref{property_1} and Property~\ref{property_2}. We parameterize the decoding distribution as a multivariate normal distribution with a non-diagonal covariance matrix:
\begin{equation}
\begin{aligned}
    & p_\theta(y_{m+1:N} \mid x_{1:N}, y_{1:m}) \\
    & = \mathcal{N}(y_{m+1:N} \mid \mu_\theta(x_{1:N}, y_{1:m}), \Sigma_\theta(x_{1:N}, y_{1:m})).
\end{aligned}
\end{equation}
Similar to TNP-D, we remove $\{(x_i, y_i)_{i=m+1}^N\}$ from the input sequence as the target points are predicted jointly. Since computing the full covariance matrix can be expensive, we consider two approximations to parameterize $\Sigma$: \\
(1) \textit{Cholesky decomposition:} $\Sigma = L L^T$, where $L$ is a lower triangular matrix with positive diagonal values; and\\
(2) \textit{Low-rank approximation:} $\Sigma = \exp(D) + A A^T$, where $D$ is a diagonal matrix and $A$ is a low-rank matrix. \\
For the main experiments of this paper, we use the Cholesky decomposition due to its computational convenience. The results for the low-rank approximation are in Appendix \ref{sec:nd_details}.

For a distribution with dimension $n$, we need a neural network that outputs $\frac{n(n+1)}{2}$ values to represent the lower triangular matrix $L$. In practice, the dimension of the distribution depends on the number of the target points, which varies during both training and evaluation. This means we cannot simply use a neural network to directly output $L$. We instead parameterize the decoder as follows. First, the output vectors $z_{m+1:N}$ of the last masked self-attention layer are fed to an MLP that produces the mean values $\mu_{m+1:N}$. We then pass $z_{m+1:N}$ through another head consisting of an additional stack of self-attention layers, and a final projection layer (an MLP)  that outputs $N-m$ vectors  $h_{m+1:N}$. Each vector $h_i \in \mathbb{R}^{p}$, is projected to a dimension $p$. The lower triangular matrix $L$ is then computed as:
\begin{equation}
    L = \text{lower}(H H^\top), \ H \in \mathbb{R}^{n \times p},
\end{equation}
where $H$ is the row-wise stack of $\{h_i\}_{i=m+1}^N$, $\text{lower}(L)$ removes the upper triangular parts of $L$, 
and $n=N-m$ is the number of the target points. 
While this particular parameterization does not universally represent all the possible lower triangular matrices, it provides two main benefits. First, it allows us to parameterize a multivariate normal distribution with an arbitrary number of target points. Second, the space complexity of this parameterization is only $\mathcal{O}(n)$ compared to $\mathcal{O}(n^2)$ if we directly output the components of $L$.

\section{Experiments}
We evaluate Transformer Neural Processes (TNPs) on several tasks: regression, image completion, Bayesian optimization, and contextual bandits. This set of experiments has been used extensively to benchmark NP-based models in prior works \citep{garnelo2018neural,kim2019attentive,lee2020bootstrapping}. We compare TNPs with other members of the NP family, namely Conditional Neural Processes (CNPs) \citep{garnelo2018conditional}, Neural Processes (NPs) \citep{garnelo2018neural}, and Bootstrapping Neural Processes (BNPs) \citep{lee2020bootstrapping}, as well as their attentive version \citep{kim2019attentive}, which are CANPs, ANPs, and BANPs, respectively. We have open-sourced the codebase for reproducing our experiments.\footnote{\url{https://github.com/tung-nd/TNP-pytorch}} The implementation of the baselines is borrowed from the official implementation of BNPs.\footnote{\url{https://github.com/juho-lee/bnp}}
\begin{table}[t]
\centering
\caption{Comparison of TNPs with the baselines on log-likelihood of the target points on various GP kernels.
We train each method with $5$ different seeds and report the mean and standard deviation.}
\label{tab:1d_benchmark}
\scalebox{0.9}{
\begin{tabular}{cccc}
\toprule
Method       & RBF            & Matérn 5/2     & Periodic   \\
\midrule
CNP          & $0.26 \pm 0.02$ & $0.04 \pm 0.02$ & $-1.40 \pm 0.02$  \\  
CANP         & $0.79 \pm 0.00$ & $0.62 \pm 0.00$ & $-7.61 \pm 0.16$   \\  
NP           & $0.27 \pm 0.01$ & $0.07 \pm 0.01$ & $-1.15 \pm 0.04$   \\  
ANP          & $0.81 \pm 0.00$ & $0.63 \pm 0.00$ & $-5.02 \pm 0.21$   \\  
BNP          & $0.38 \pm 0.02$ & $0.18 \pm 0.02$ & $\boldsymbol{-0.96 \pm 0.02}$ \\  
BANP         & $0.82 \pm 0.01$ & $0.66 \pm 0.00$ & $-3.09 \pm 0.14$ \\  
TNP-D         & $1.39 \pm 0.00$  & $0.95 \pm 0.01$  & $-3.53 \pm 0.37$    \\
TNP-A          & $\boldsymbol{1.63 \pm 0.00}$  & $\boldsymbol{1.21 \pm 0.00}$  & $-2.26 \pm 0.17$  \\
TNP-ND & $1.46 \pm 0.00$  & $1.02 \pm 0.00$  & $-4.13 \pm 0.33$   \\
\bottomrule
\end{tabular}
}
\vspace{-0.1in}
\end{table}
\subsection{1-D Regression} \label{sec:1d_regression_compare}

\textbf{Problem}: Given a set of context points $\{x_i,y_i\}_{i=1}^m$ that come from an unknown function $f$, we train the model 
to make predictions for a set of target points $\{x_i\}_{i=m+1}^N$ that come from the same function \citep{garnelo2018neural}.

\textbf{Training}: In each epoch of training, we draw $B$ different functions from a Gaussian Process prior with an RBF kernel: $f_i \sim \mathcal{G}\mathcal{P}(m, k)$, where $m(x) = 0$ and $k(x,x') = \sigma_f^2 \exp(-\frac{(x-x')^2}{2\ell^2})$. The hyperparameters of the GP ($\ell$ and $\sigma_f$) are randomized for each function, allowing the model to learn from a more diverse set of functions. For each $f_i$, we choose $N$ random locations to evaluate, and sample an index $m$ that splits the sequence to context and target points. For all methods, $\ell \sim \mathcal{U}[0.6, 1.0), \sigma_f \sim \mathcal{U}[0.1, 1.0)$, $B = 16$, $N \sim \mathcal{U}[6,50)$, $m \sim \mathcal{U}[3, 47)$.

\textbf{Evaluation}: We test the trained models on unseen functions that are drawn from GPs with RBF, Matérn 5/2 
and Periodic kernels. The number of evaluation points $N$ and the number of context points $m$ are generated from the same uniform distribution as in training. The evaluation set contains $48000$ functions for each kernel. 
We evaluate all methods on log-likelihood of the target points.
We refer the readers to Appendix \ref{sec:1d_additional} 
for more evaluation metrics.

\textbf{Results}: Table \ref{tab:1d_benchmark} shows that TNPs outperform the other methods on 2/3 kernels by a large margin.
Even though TNPs underperform on regressing functions from the periodic kernel, we show later in Section~\ref{sec:bo} that they can optimize the same class of functions better than the baselines.
As expected, among three variants of TNPs, TNP-A achieves the best likelihood due to the use of an autoregressive decoder, followed by TNP-ND, and finally TNP-D.

\begin{table*}[t]
\centering
\caption{Comparison of TNPs with the baselines on log-likelihood of the target points on two datasets: EMNIST (left) and CelebA (right). We train each method with $5$ different seeds and report the mean and standard deviation.}
\label{tab:img_benchmark}
\scalebox{0.9}{
\begin{tabular}{cc}
\toprule
\multirow{2}{*}{Method}       & \multirow{2}{*}{CelebA} \\
\\
\midrule
CNP          & $2.15 \pm 0.01$ \\
CANP         & $2.66 \pm 0.01$ \\
NP           & $2.48 \pm 0.02$ \\
ANP          & $2.90 \pm 0.00$ \\
BNP          & $2.76 \pm 0.01$ \\
BANP         & $3.09 \pm 0.00$ \\
TNP-D         & $3.89 \pm 0.01$ \\ 
TNP-A          & $\boldsymbol{5.82 \pm 0.01}$  \\
TNP-ND & $5.48 \pm 0.02$  \\
\bottomrule
\end{tabular}
}
\quad
\quad
\scalebox{0.9}{
\begin{tabular}{ccc}
\toprule
\multirow{2}{*}{Method} & \multicolumn{2}{c}{EMNIST} \\
                        & Seen classes (0-9) & Unseen classes (10-46) \\
\midrule
CNP          & $0.73 \pm 0.00$     & $0.49 \pm 0.01$ \\
CANP         & $0.94 \pm 0.01$     & $0.82 \pm 0.01$ \\
NP           & $0.79 \pm 0.01$     & $0.59 \pm 0.01$ \\
ANP          & $0.98 \pm 0.00$     & $0.89 \pm 0.00$ \\
BNP          & $0.88 \pm 0.01$     & $0.73 \pm 0.01$ \\
BANP         & $1.01 \pm 0.00$     & $0.94 \pm 0.00$ \\
TNP-D        & $1.46 \pm 0.01$     & $1.31 \pm 0.00$ \\
TNP-A        & $\boldsymbol{1.54 \pm 0.01}$     & $\boldsymbol{1.41 \pm 0.01}$ \\
TNP-ND       & $1.50 \pm 0.00$     & $1.31 \pm 0.00$ \\
\bottomrule
\end{tabular}
}
\end{table*}

\begin{figure}[t]
     \centering
     \includegraphics[width=0.7\linewidth]{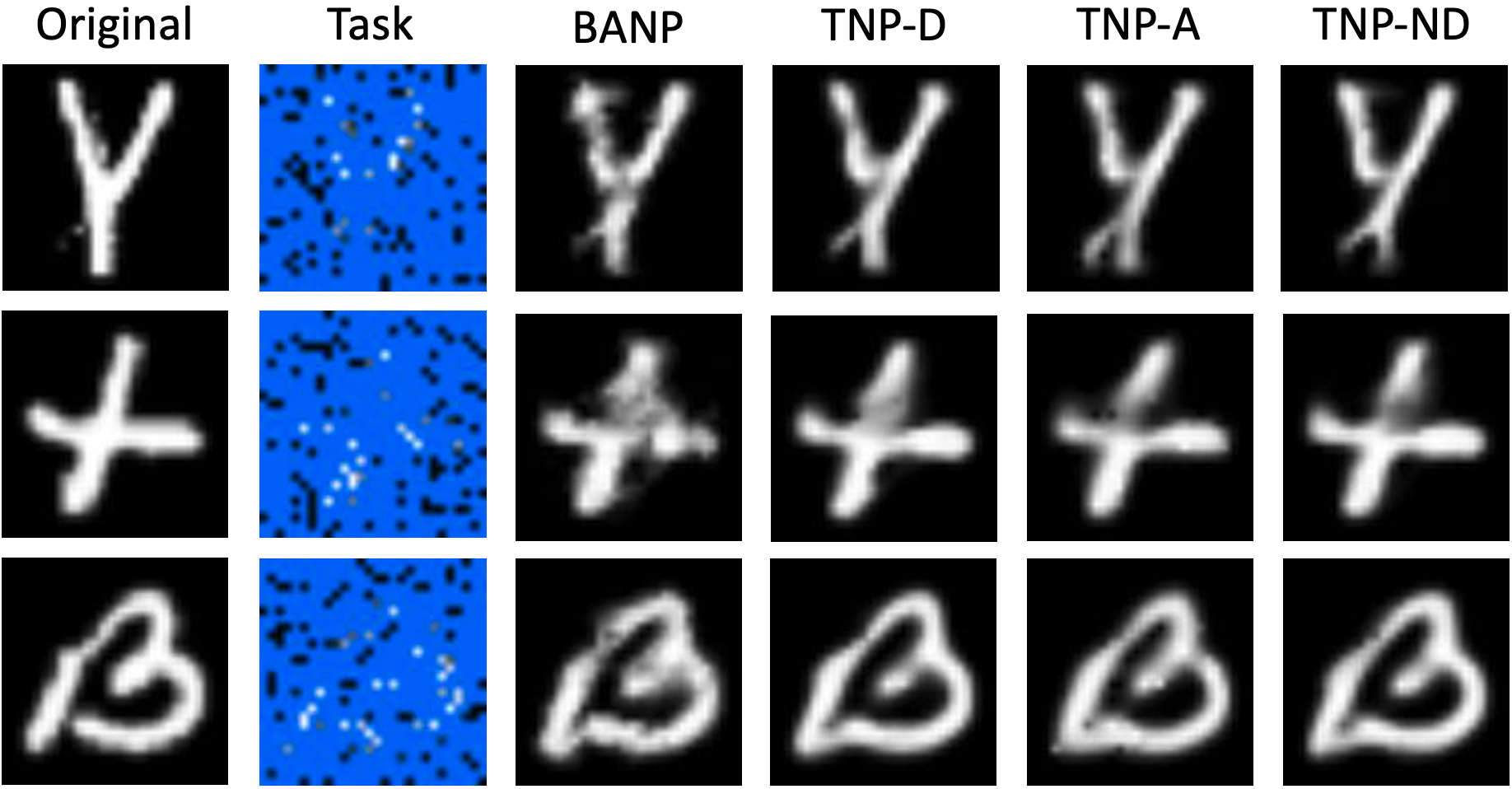}
     \caption{Completed images produced by the best baseline and TNPs from $100$ context points. Original images are drawn randomly from EMNIST unseen classes. For stochastic models, we sample multiple times and average the results.}
     \label{fig:emnist_main}
     \vspace{-0.2in}
\end{figure}
\subsection{Image completion}
\textbf{Problem}: The model observes a subset of pixel values of an image and completes the rest. This can be cast as a 2-D meta-regression problem, in which $x$ denotes the coordinates of a pixel, and $y$ denotes the corresponding pixel value. Each image can be thought of as a unique function that maps from coordinate to pixel intensity \citep{garnelo2018neural}.

\textbf{Training}: We use two datasets for this experiment: 
EMNIST \citep{cohen2017emnist} and CelebA \citep{liu2018large}. EMNIST contains black and white images of handwritten letters, and CelebA contains colored 
images of celebrity faces. We down-sample each image to $32 \times 32$. For EMNIST, we only use $10$ classes for training. Similar to the 1-D regression experiment, we randomly select subsets of pixels as context points and target points. 
For both datasets, $N \sim \mathcal{U}[6, 200), m \sim \mathcal{U}[3, 197)$. The $x$ values are rescaled to $[-1, 1]$ and the $y$ values are rescaled to $[-0.5, 0.5]$.

\textbf{Evaluation}: We evaluate each method on log-likelihood of the target points on held-out datasets. The number of pixels and the number of context points are generated from the same uniform distribution as in training.

\textbf{Results}: Table \ref{tab:img_benchmark} show significant improvements of TNPs over the baselines. Similar to the 1-D regression, TNP-A achieves the best likelihood, followed by TNP-ND and TNP-D. For EMNIST, the performance of TNPs on unseen classes is only slightly worse than on seen classes, indicating better generalization compared to the other methods. Figure \ref{fig:emnist_main} shows that TNPs produce noticeably better completed images than the best baseline. We refer the readers to Appendix \ref{image_add} for different samples produced by TNPs.

\begin{figure}[t]
    \centering
    \includegraphics[width=0.7\linewidth]{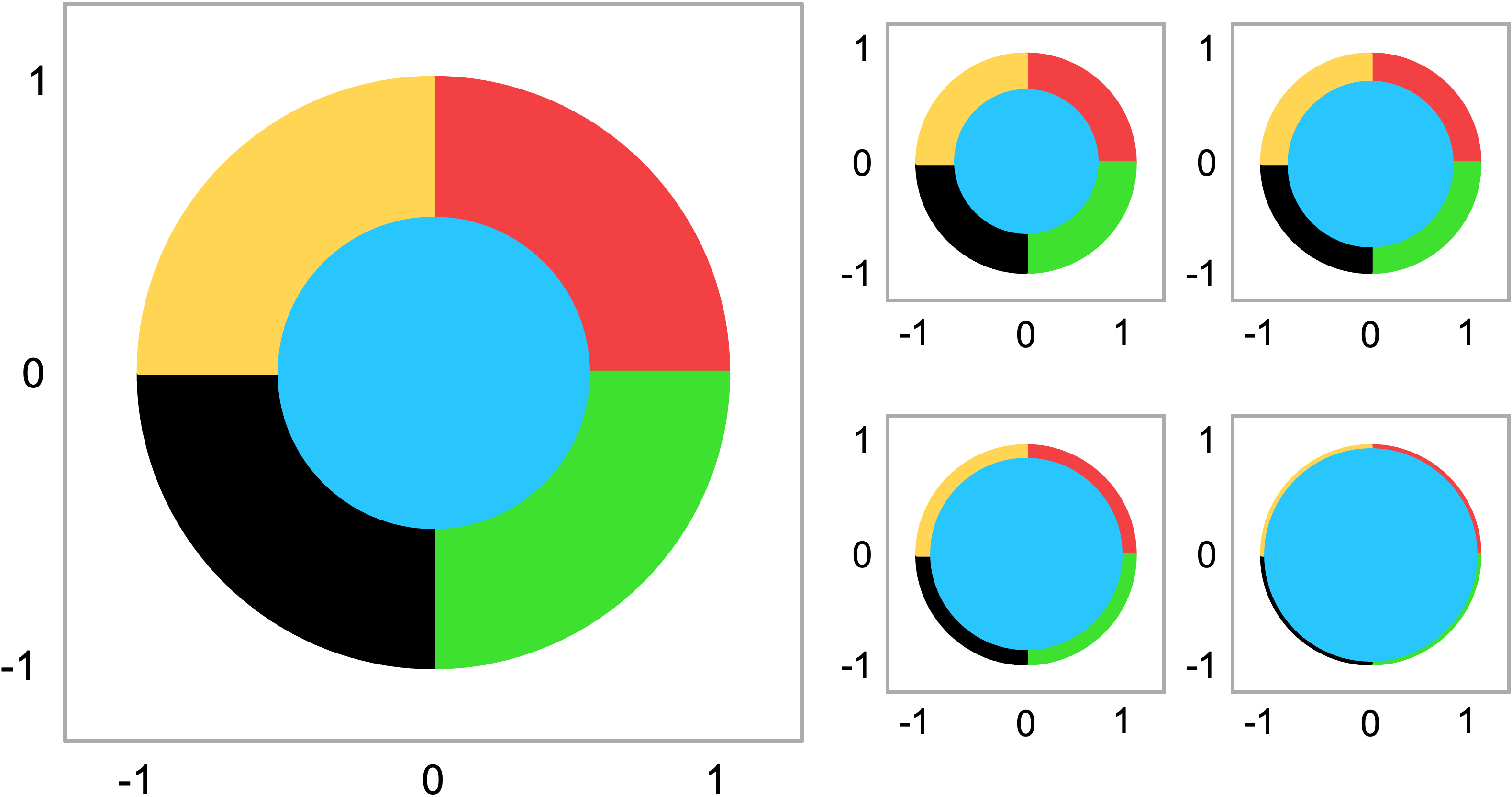}
    \caption{The wheel bandit problem with varying values of $\delta$. \vspace{-0.3cm}}
    \label{fig:wheel_bandit}
\end{figure}
\begin{table*}[t]
\centering
\caption{Comparison of TNPs with the baselines on cumulative regret on contextual bandit problems with different values of $\delta$. We run each model $50$ times for each value of $\delta$ and report the mean and standard deviation.}
\label{tab:contextual_benchmark}
\scalebox{0.9}{
\begin{tabular}{cccccccc}
\toprule
Method       & $\delta=0.7$   & $\delta=0.9$   & $\delta=0.95$  & $\delta=0.99$  & $\delta=0.995$  & $\delta=0.999$  & Average  \\ 
\midrule
Uniform      & $100.00 \pm 1.18$  & $100.00 \pm 3.03$  & $100.00 \pm 4.16$  & $100.00 \pm 7.52$  & $100.00 \pm 8.11$   & $100.00 \pm 7.96$  & $100.00 \pm 5.97$   \\ 
CNP          & $4.08 \pm 0.29$    & $8.14 \pm 0.33$    & $8.01 \pm 0.40$    & $26.78 \pm 0.85$   & $38.25 \pm 1.01$    & $93.17 \pm 2.81$   & $29.74 \pm 30.85$    \\ 
CANP         & $8.08 \pm 9.93$    & $11.69 \pm 11.96$  & $24.49 \pm 13.25$  & $47.33 \pm 20.49$  & $49.59 \pm 17.87$   & $33.29 \pm 5.05$   & $29.08 \pm 21.28$    \\ 
NP           & $1.56 \pm 0.13$    & $2.96 \pm 0.28$    & $4.24 \pm 0.22$    & $18.00 \pm 0.42$   & $25.53 \pm 0.18$    & $62.73 \pm 1.49$   & $19.17 \pm 21.36$    \\ 
ANP          & $1.62 \pm 0.16$    & $4.05 \pm 0.31$    & $5.39 \pm 0.50$    & $19.57 \pm 0.67$   & $27.65 \pm 0.95$    & $73.36 \pm 5.95$   & $21.94 \pm 24.92$    \\ 
BNP          & $62.51 \pm 1.07$   & $57.49 \pm 2.13$   & $58.22 \pm 2.27$   & $58.91 \pm 3.77$   & $62.50 \pm 4.85$    & $77.46 \pm 6.18$   & $62.85 \pm 7.78$    \\ 
BANP         & $4.23 \pm 16.58$   & $12.42 \pm 29.58$  & $31.10 \pm 36.10$  & $52.59 \pm 18.11$  & $49.55 \pm 14.52$   & $45.45 \pm 11.71$  & $32.56 \pm 29.43$   \\ 
TNP-D & $\boldsymbol{1.18 \pm 0.94}$    & $1.70 \pm 0.41$    & $2.55 \pm 0.43$    & $\boldsymbol{3.57 \pm 1.22}$    & $\boldsymbol{4.68 \pm 1.09}$     & $9.56 \pm 0.44$    & $\boldsymbol{3.87 \pm 2.91}$     \\
TNP-A          & $3.67 \pm 4.88$ & $4.04 \pm 2.38$ & $4.29 \pm 2.36$ & $5.79 \pm 5.27$ & $9.29 \pm 7.62$ & $\boldsymbol{6.13 \pm 2.50}$ & $5.54 \pm 4.98$ \\
TNP-ND & $1.76 \pm 0.61$ & $\boldsymbol{1.41 \pm 0.98}$ & $\boldsymbol{1.61 \pm 1.65}$ & $4.98 \pm 2.84$ & $7.22 \pm 3.28$ & $13.66 \pm 2.92$ & $5.11 \pm 4.94$ \\
\bottomrule
\end{tabular}
}
\end{table*}
\subsection{Contextual bandits}

\textbf{Problem}: We compare TNPs with the baselines on the wheel bandit problem introduced in \citet{riquelme2018deep} (Figure \ref{fig:wheel_bandit}). In this problem, a unit circle is divided into a low-reward region (blue area) and four high-reward regions (the other four coloured areas). A scalar $\delta$ determines the size of the low-reward region, and other regions have equal sizes. The agent does not know the underlying $\delta$, and has to choose among $k = 5$ arms given its coordinates $X = (X_1, X_2)$ within the circle. If $||X|| \leq \delta$, the agent falls within the low-reward region (blue). In this case the optimal action is $k = 1$, which provides a reward $r \sim \mathcal{N}(1.2, 0.012)$, while all other actions only return $r \sim \mathcal{N}(1.0, 0.012)$. If the agent falls within any of the four high-reward region $(||X|| > \delta)$, the optimal arm will be one of the remaining four $k = 2-5$, depending on the specific area. Pulling the optimal arm here results in a high reward $r \sim \mathcal{N}(50.0, 0.012)$, and as before all other arms receive $\mathcal{N}(1.0, 0.012)$ except for arm $k = 1$ which always returns $\mathcal{N}(1.2, 0.012)$.

\textbf{Training}: We sample a dataset of $B$ different wheel problems $\{\delta_i\}_{i=1}^B$, which are drawn from a uniform distribution $\delta \sim \mathcal{U}(0,1)$. For each problem, we sample $N$ points to evaluate and pick $m$ points as context, in which each point is a tuple $(X,r)$ of the coordinates $X$ and the corresponding reward values $r$ of all $5$ arms. The training objective is to regress the reward values from the coordinates. We set $B=8, N=562, m=512$ in our experiments.

\textbf{Evaluation}: We test TNPs and the baselines on problems with varying $\delta$ values. We run with $50$ different seeds for each value of $\delta$, and each run consists of $2000$ steps. In each step, the agent predicts the reward values of $5$ arms based on the coordinates $X$, chooses an arm according to the Upper Confidence Bound (UCB) algorithm and receives the ground-truth reward value for the chosen arm. We use cumulative regret as the evaluation metric. See Appendix \ref{sec:bandit_add} for simple regret results.

\begin{figure*}[t]
    \centering
    \includegraphics[width=0.7\textwidth]{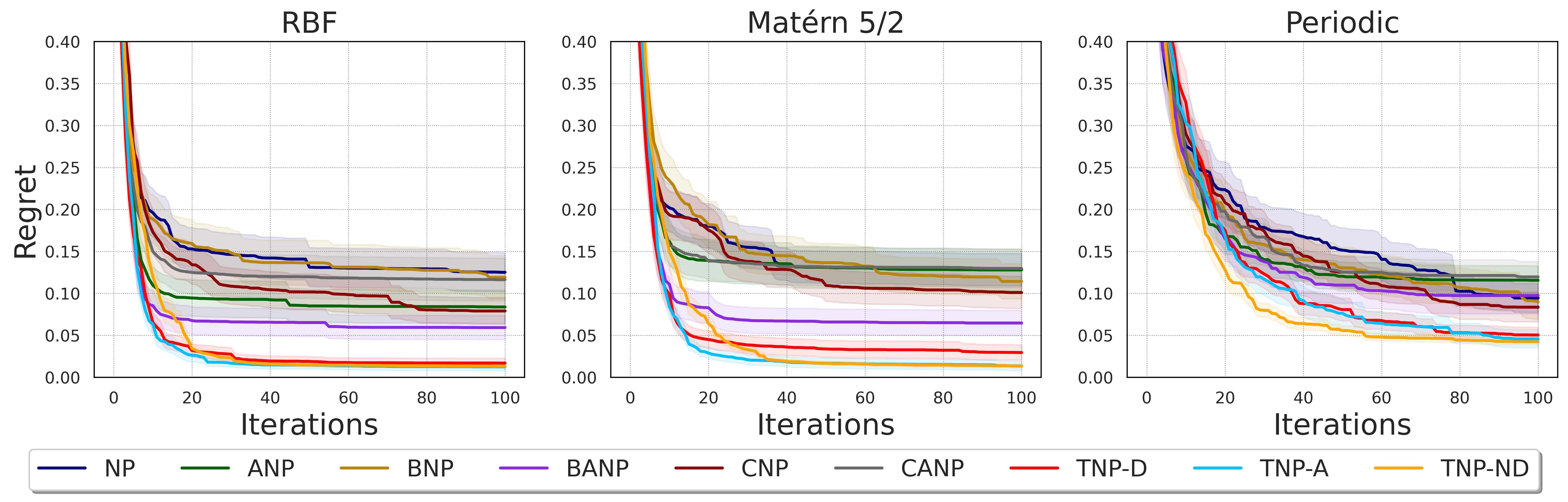}
    \caption{Regret performance on 1D BO tasks. For each kernel, we generate $100$ functions and report the mean and standard deviation.}
    \label{fig:bo_1d}
\end{figure*} 
\begin{figure*}[t!]
    \centering
    \includegraphics[width=0.7\textwidth]{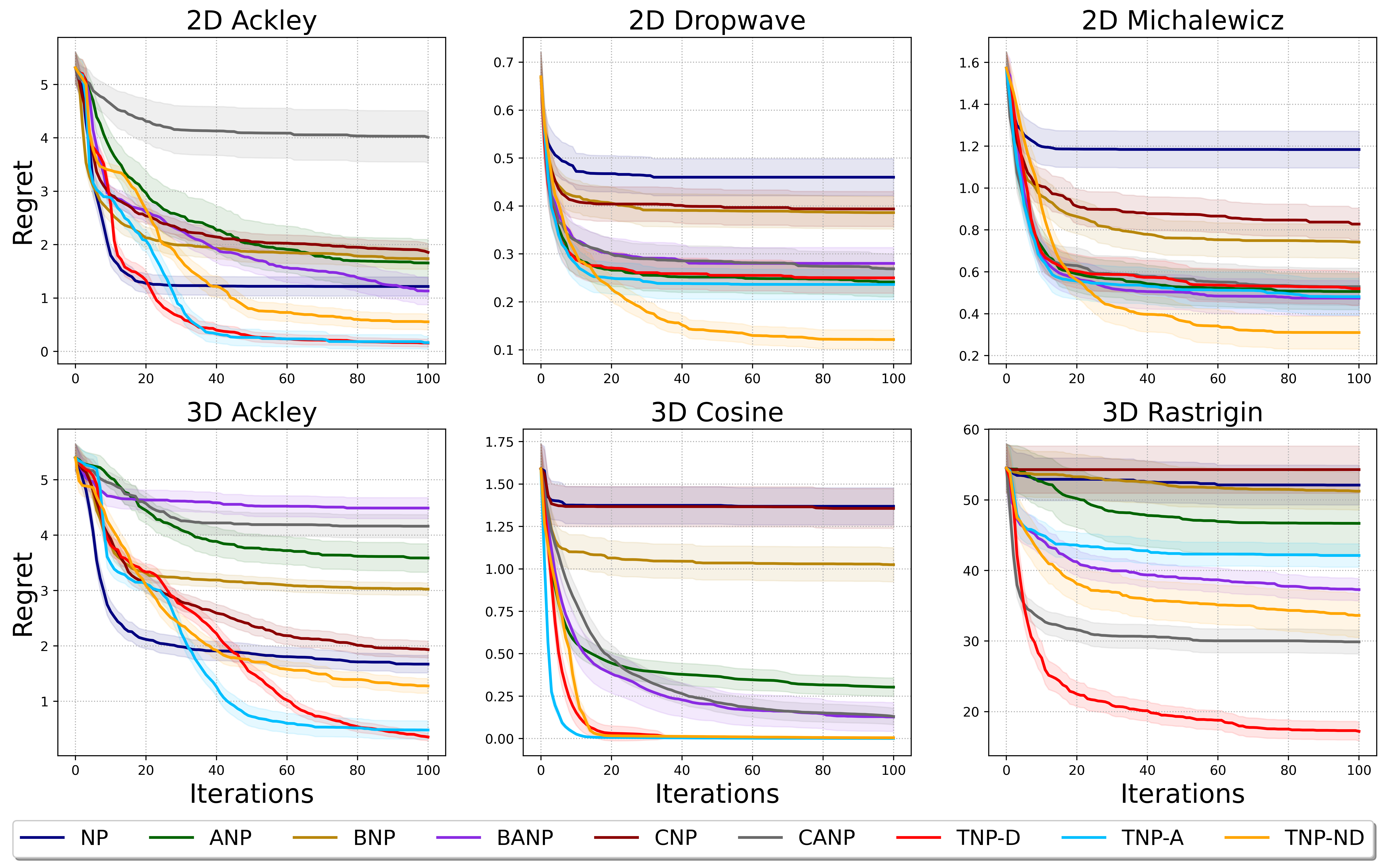}
    \caption{Regret performance on 2D and 3D BO tasks. For each function, we run BO for $100$ times with different seeds and report the mean and standard deviation.}
    \label{fig:bo_multi}
\end{figure*}
\textbf{Results}: Table \ref{tab:contextual_benchmark} shows that TNPs outperform all baselines by a large margin on all settings, especially for harder problems (higher values of $\delta$). Moreover, the performance of Transformers only slightly drops when the difficulty increases, which is a significant improvement over the baselines, as these methods barely work for hard problems. We also note that for 
other NP-based methods, using attention 
hurts the performance. This can be clearly seen from the table, as CANPs, ANPs and BANPs are inferior to the non-attentive counterparts CNPs, NPs, and BNPs, respectively.

\subsection{Bayesian Optimization}\label{sec:bo}
\textbf{Problem}: The goal of Bayesian optimization (BO) \citep{frazier2018tutorial} is to optimize a black-box function $f(x)$ that we can evaluate but do not have access to the gradient information. The BO process runs in a loop, where in each iteration, we use a surrogate function to estimate the unknown function $f$, and an acquisition function to choose the next point to evaluate based on the estimation. We use TNPs and the baselines as the surrogate function, and UCB as the acquisition function. We consider both one-dimensional (1D) and multi-dimensional (2D and 3D) settings.

\textbf{Training}: Training in the one-dimensional setting is identical to training in Section \ref{sec:1d_regression_compare}. When $x$ is multi-dimensional, we use GPytorch \citep{gardner2018gpytorch} to generate training data from multivariate GPs with RBF kernel. We use $N \sim \mathcal{U}[60,128), m \sim \mathcal{U}[30, 98)$ for the 2D setting, and $N \sim \mathcal{U}[128,256), m \sim \mathcal{U}[64, 192)$ for the 3D setting.

\textbf{Evaluation}: In the 1D setting, the objective functions are generated from GPs with RBF, Matérn 5/2, and Periodic kernels. In the multi-dimensional setting, we use various benchmark functions in the optimization literature \citep{KimJ2017bayeso,KimJ2020benchmark}, and Botorch \citep{balandat2020botorch} is used as the implementation of the overall BO process (e.g., optimization and acquisition function). For each objective function, we run BO for $100$ iterations, and simple regret is used as the evaluation metric.

\textbf{Results}: Figure \ref{fig:bo_1d} shows that TNPs outperform the baselines significantly in all three kernels. This includes the Periodic kernel, where TNPs were outperformed by other methods for meta regression (Table \ref{tab:1d_benchmark}), suggesting that simple meta regression metrics may not be sufficient for model selection in online decision making.
Figure \ref{fig:bo_multi} shows the results for the multi-dimensional setting. All three TNP variants outperform the baselines in at least $3/6$ tasks: 2D Ackley, 3D Ackley, and 3D Cosine, and achieve competitive results to the best baseline in 2D Dropwave and 2D Michalewicz. TNP-ND is the best variant in this setting, which beats the baselines in $5/6$ tasks, followed by TNP-D, which outperforms the baselines in $4/6$ tasks. TNP-A also performs well and is competitive on all but the 3D Rastrigin task.

\subsection{Memory and time complexity of TNPs}
The major successes of Transformers in language~\cite{brown2020language} and vision~\cite{dosovitskiy2020image} largely attribute to the ability to scale to billions of parameters, thus having unprecedented results. Similarly, one can claim that the superior performance of TNPs is merely due to having more parameters than the baselines. Therefore, we compare the parameter counts, training and prediction time of TNPs and the baselines in Table \ref{tab:time_memory}. All models are comparable, barring TNP-A which as expected, incurs a higher prediction time because of the autoregressive nature. This indicates that scaling is not all we need, but the transformer architecture itself is important too.
\begin{table}[t]
\centering
\caption{Number of parameters, training time and prediction time of TNPs and the baselines. We measure run time on the 1-D regression task on an RTX2080Ti, with $1000$ batches of size $16$.}
\label{tab:time_memory}
\begin{tabular}{@{}lccc@{}}
\toprule
Method  & \# Parameters & Training (s) & Prediction (s) \\ \midrule
CNP    & $215682$      & $11.20$           & $0.76$              \\
CANP   & $331906$      & $20.05$           & $1.64$              \\
NP     & $232194$      & $16.79$           & $1.24$              \\
ANP    & $348418$      & $21.07$           & $2.15$              \\
BNP    & $248450$      & $20.07$           & $4.86$              \\
BANP   & $364674$      & $24.07$           & $16.72$             \\
TNP-A  & $222082$      & $20.14$           & $337.09$            \\
TNP-D  & $222082$      & $20.13$           & $2.75$              \\
TNP-ND & $332821$      & $26.89$           & $4.05$              \\ \bottomrule
\end{tabular}

\end{table}

\section{Related Work} \label{sec:related}
\paragraph{Neural Processes.} The Conditional Neural Process (CNP) \citep{garnelo2018conditional} was the first member of the NP family. 
CNPs encode the context points to a deterministic latent vector $z$ and hence do not have a notion of functional uncertainty.
The Neural Process (NP) \citep{garnelo2018neural} was proposed to address these shortcomings by introducing stochasticity in the form of latent variables. \citet{le2018empirical} made a careful empirical examination of different objectives and hyperparameter choices when training a Neural Process. 
Since then, many extensions have been proposed which improve NPs by building translation equivariance \citep{gordon2019convolutional}, incorporating attention to address underfitting \citep{kim2019attentive}, using bootstrapping to overcome the limitations of Gaussian assumption on the latent variables \citep{lee2020bootstrapping}, and modeling predictive correlations \citep{bruinsma2021gaussian}. In addition, many works have also extended NPs to a wider range of problems, which include modeling a sequence of stochastic processes \citep{singh2019sequential} and modeling stochastic physics fields \citep{holderrieth2021equivariant}. \citet{mathieu2021contrastive} recently proposed a contrastive learning framework that replaces the reconstruction objective in NP to learn a better representation. In \citet{xu2020metafun}, the authors proposed a meta-learner based on CANP that achieved good results on large scale image classification. \citet{groverprobing} provide a series of fine-grained diagnostics for the uncertainty modeled via various neural processes. Finally, \citet{galashov2019meta} empirically studied the effectiveness of NPs as a meta-learner in various sequential decision making problems.

\paragraph{Transformers.} The transformer
architecture was introduced by \citet{vaswani2017attention} for flexible modeling of natural language. 
In recent years, transformers have made breakthroughs in 
language \citep{devlin2018bert,radford2018improving,radford2019language,brown2020language} and vision \citep{dosovitskiy2020image,radford2021learning,chen2020generative}. Recent works \citep{lee2019set,kossen2021self} have also applied transformers to modeling the relationship between different data points, in addition to modeling interactions between components of the same data point. 
Set Transformers \citep{lee2019set} are closely related to TNPs, 
and were proposed to solve set-input problems, such as finding the maximum number in a set or counting unique characters. Set Transformers adopt the transformer architecture to model the interactions between items of the same set and remove positional encodings for permutation invariance. However, our focus in this work is on uncertainty-aware meta learning.
For such tasks, TNPs incorporate important architectural design choices such as the input representation and masking mechanism. 
It is crucial for TNPs to model the interaction between the inputs $x's$ as well as the relationship between $x$ and $y$ to accurately infer the underlying function. \citet{chen2021decision} proposed an autoregressive transformer-based model for offline reinforcement learning, which was extended to the online setting by \citet{onlinedt}. Besides differences in the agent setup, this work considers trajectories from a single task, unlike our focus on meta-learning.
Concurrently,~\citet{mullertransformers} developed Prior-Data Fitted Networks (PFNs), a method very similar to TNPs. PFNs can make predictions for novel tasks in a single forward propagation and satisfies the two desired properties~\ref{property_1} and~\ref{property_2}. However, PFNs cast meta-learning as a Bayesian inference problem, wherein the model learns to approximate the posterior distribution of the ground truth given the input and the training data. The training data thus corresponds to the context points in TNPs, while the input and ground truth correspond to the target points. PFNs focus on the Bayesian inference aspect, while we are more interested in uncertainty-aware tasks, leading to different experimental settings. Moreover, PFNs only predict the marginal distributions, which are equivalent to TNP-D. 

\section{Discussion}

Neural Processes offer a promising approach for learning flexible stochastic processes directly from data.
However, it is unclear which aspects of their design are essential for downstream applications in uncertainty-aware meta learning, such as regression and sequential decision making.
Popular claims, such as the use of latent variables for representing functional uncertainty and diverse sampling, have mixed empirical evidence~\cite{garnelo2018neural} and a deterministic path between the encoder and decoder is important for good performance~\cite{le2018empirical}.
Moreover, the standard evidence lower bounds for variational autoencoders (including NPs) can completely ignore the latent code with powerful decoders~\cite{chen2016variational,alemi2018fixing}.

In this regard, we proposed Transformer Neural Processes (TNPs), an alternative framing of uncertainty-aware meta learning via sequence modeling.
TNPs optimize an autoregressive modeling objective and benefit from the use of a transformer backbone~\cite{vaswani2017attention,radford2018improving}.
While attention mechanisms have also previously been used for parameterizing NPs~\cite{kim2019attentive}, we showed that replacing the entire architecture stack can drastically improve performance across various benchmark tasks.
We also showed these empirical benefits cannot be attributed solely to the use of a more expressive decoding distribution as in TNP-A, but can also be obtained to a good degree via tractable and equivariant (but relatively less expressive) parameterizations such as TNP-D and TNP-ND.

In the future, we are keen to further build on the merits of the TNP architecture for scaling to high-dimensional problems beyond current benchmarks. We are also interested in pursuing future work towards a clean separation of functional and point uncertainties (as in GPs), potentially via recent advances in stochastic transformers~\cite{lin2020variational}.

\section*{Acknowledgements}
We would like to thank Hritik Bansal, Shashank Goel, Siddarth Krishnamoorthy, Satvik Mashkaria, and Tuan Pham for the insightful discussions during the early development of the paper. We also want to thank the Google TPU Research Cloud program for the computing resources.

\bibliography{main}

\begin{thebibliography}{50}
\providecommand{\natexlab}[1]{#1}
\providecommand{\url}[1]{\texttt{#1}}
\expandafter\ifx\csname urlstyle\endcsname\relax
  \providecommand{\doi}[1]{doi: #1}\else
  \providecommand{\doi}{doi: \begingroup \urlstyle{rm}\Url}\fi

\bibitem[Alemi et~al.(2018)Alemi, Poole, Fischer, Dillon, Saurous, and
  Murphy]{alemi2018fixing}
Alemi, A., Poole, B., Fischer, I., Dillon, J., Saurous, R.~A., and Murphy, K.
\newblock Fixing a broken elbo.
\newblock In \emph{International Conference on Machine Learning}, pp.\
  159--168. PMLR, 2018.

\bibitem[Balandat et~al.(2020)Balandat, Karrer, Jiang, Daulton, Letham, Wilson,
  and Bakshy]{balandat2020botorch}
Balandat, M., Karrer, B., Jiang, D., Daulton, S., Letham, B., Wilson, A.~G.,
  and Bakshy, E.
\newblock Botorch: A framework for efficient monte-carlo bayesian optimization.
\newblock \emph{Advances in neural information processing systems}, 33, 2020.

\bibitem[Brown et~al.(2020)Brown, Mann, Ryder, Subbiah, Kaplan, Dhariwal,
  Neelakantan, Shyam, Sastry, Askell, et~al.]{brown2020language}
Brown, T.~B., Mann, B., Ryder, N., Subbiah, M., Kaplan, J., Dhariwal, P.,
  Neelakantan, A., Shyam, P., Sastry, G., Askell, A., et~al.
\newblock Language models are few-shot learners.
\newblock \emph{arXiv preprint arXiv:2005.14165}, 2020.

\bibitem[Bruinsma et~al.(2021)Bruinsma, Requeima, Foong, Gordon, and
  Turner]{bruinsma2021gaussian}
Bruinsma, W.~P., Requeima, J., Foong, A.~Y., Gordon, J., and Turner, R.~E.
\newblock The gaussian neural process.
\newblock \emph{arXiv preprint arXiv:2101.03606}, 2021.

\bibitem[Cesa-Bianchi \& Lugosi(2006)Cesa-Bianchi and
  Lugosi]{cesa2006prediction}
Cesa-Bianchi, N. and Lugosi, G.
\newblock \emph{Prediction, learning, and games}.
\newblock Cambridge university press, 2006.

\bibitem[Chan et~al.(2022)Chan, Santoro, Lampinen, Wang, Singh, Richemond,
  McClelland, DeepMind, and Hill]{chan2022data}
Chan, S.~C., Santoro, A., Lampinen, A.~K., Wang, J.~X., Singh, A., Richemond,
  P.~H., McClelland, J., DeepMind, S., and Hill, F.
\newblock Data distributional properties drive emergent in-context learning in
  transformers.
\newblock \emph{CoRR}, 2022.

\bibitem[Chen et~al.(2021)Chen, Lu, Rajeswaran, Lee, Grover, Laskin, Abbeel,
  Srinivas, and Mordatch]{chen2021decision}
Chen, L., Lu, K., Rajeswaran, A., Lee, K., Grover, A., Laskin, M., Abbeel, P.,
  Srinivas, A., and Mordatch, I.
\newblock Decision transformer: Reinforcement learning via sequence modeling.
\newblock \emph{arXiv preprint arXiv:2106.01345}, 2021.

\bibitem[Chen et~al.(2020)Chen, Radford, Child, Wu, Jun, Luan, and
  Sutskever]{chen2020generative}
Chen, M., Radford, A., Child, R., Wu, J., Jun, H., Luan, D., and Sutskever, I.
\newblock Generative pretraining from pixels.
\newblock In \emph{International Conference on Machine Learning}, pp.\
  1691--1703. PMLR, 2020.

\bibitem[Chen et~al.(2016)Chen, Kingma, Salimans, Duan, Dhariwal, Schulman,
  Sutskever, and Abbeel]{chen2016variational}
Chen, X., Kingma, D.~P., Salimans, T., Duan, Y., Dhariwal, P., Schulman, J.,
  Sutskever, I., and Abbeel, P.
\newblock Variational lossy autoencoder.
\newblock \emph{arXiv preprint arXiv:1611.02731}, 2016.

\bibitem[Chung et~al.(2021)Chung, Char, Guo, Schneider, and
  Neiswanger]{chung2021uncertainty}
Chung, Y., Char, I., Guo, H., Schneider, J., and Neiswanger, W.
\newblock Uncertainty toolbox: an open-source library for assessing,
  visualizing, and improving uncertainty quantification.
\newblock \emph{arXiv preprint arXiv:2109.10254}, 2021.

\bibitem[Cohen et~al.(2017)Cohen, Afshar, Tapson, and
  Van~Schaik]{cohen2017emnist}
Cohen, G., Afshar, S., Tapson, J., and Van~Schaik, A.
\newblock Emnist: Extending mnist to handwritten letters.
\newblock In \emph{2017 International Joint Conference on Neural Networks
  (IJCNN)}, pp.\  2921--2926. IEEE, 2017.

\bibitem[Devlin et~al.(2018)Devlin, Chang, Lee, and Toutanova]{devlin2018bert}
Devlin, J., Chang, M.-W., Lee, K., and Toutanova, K.
\newblock Bert: Pre-training of deep bidirectional transformers for language
  understanding.
\newblock \emph{arXiv preprint arXiv:1810.04805}, 2018.

\bibitem[Dosovitskiy et~al.(2020)Dosovitskiy, Beyer, Kolesnikov, Weissenborn,
  Zhai, Unterthiner, Dehghani, Minderer, Heigold, Gelly,
  et~al.]{dosovitskiy2020image}
Dosovitskiy, A., Beyer, L., Kolesnikov, A., Weissenborn, D., Zhai, X.,
  Unterthiner, T., Dehghani, M., Minderer, M., Heigold, G., Gelly, S., et~al.
\newblock An image is worth 16x16 words: Transformers for image recognition at
  scale.
\newblock \emph{arXiv preprint arXiv:2010.11929}, 2020.

\bibitem[Foong et~al.(2020)Foong, Bruinsma, Gordon, Dubois, Requeima, and
  Turner]{foong2020meta}
Foong, A.~Y., Bruinsma, W.~P., Gordon, J., Dubois, Y., Requeima, J., and
  Turner, R.~E.
\newblock Meta-learning stationary stochastic process prediction with
  convolutional neural processes.
\newblock \emph{arXiv preprint arXiv:2007.01332}, 2020.

\bibitem[Frazier(2018)]{frazier2018tutorial}
Frazier, P.~I.
\newblock A tutorial on bayesian optimization.
\newblock \emph{arXiv preprint arXiv:1807.02811}, 2018.

\bibitem[Galashov et~al.(2019)Galashov, Schwarz, Kim, Garnelo, Saxton, Kohli,
  Eslami, and Teh]{galashov2019meta}
Galashov, A., Schwarz, J., Kim, H., Garnelo, M., Saxton, D., Kohli, P., Eslami,
  S., and Teh, Y.~W.
\newblock Meta-learning surrogate models for sequential decision making.
\newblock \emph{arXiv preprint arXiv:1903.11907}, 2019.

\bibitem[Gardner et~al.(2018)Gardner, Pleiss, Bindel, Weinberger, and
  Wilson]{gardner2018gpytorch}
Gardner, J.~R., Pleiss, G., Bindel, D., Weinberger, K.~Q., and Wilson, A.~G.
\newblock Gpytorch: Blackbox matrix-matrix gaussian process inference with gpu
  acceleration.
\newblock \emph{arXiv preprint arXiv:1809.11165}, 2018.

\bibitem[Garnelo et~al.(2018{\natexlab{a}})Garnelo, Rosenbaum, Maddison,
  Ramalho, Saxton, Shanahan, Teh, Rezende, and Eslami]{garnelo2018conditional}
Garnelo, M., Rosenbaum, D., Maddison, C., Ramalho, T., Saxton, D., Shanahan,
  M., Teh, Y.~W., Rezende, D., and Eslami, S.~A.
\newblock Conditional neural processes.
\newblock In \emph{International Conference on Machine Learning}, pp.\
  1704--1713. PMLR, 2018{\natexlab{a}}.

\bibitem[Garnelo et~al.(2018{\natexlab{b}})Garnelo, Schwarz, Rosenbaum, Viola,
  Rezende, Eslami, and Teh]{garnelo2018neural}
Garnelo, M., Schwarz, J., Rosenbaum, D., Viola, F., Rezende, D.~J., Eslami, S.,
  and Teh, Y.~W.
\newblock Neural processes.
\newblock \emph{arXiv preprint arXiv:1807.01622}, 2018{\natexlab{b}}.

\bibitem[Gordon et~al.(2019)Gordon, Bruinsma, Foong, Requeima, Dubois, and
  Turner]{gordon2019convolutional}
Gordon, J., Bruinsma, W.~P., Foong, A.~Y., Requeima, J., Dubois, Y., and
  Turner, R.~E.
\newblock Convolutional conditional neural processes.
\newblock \emph{arXiv preprint arXiv:1910.13556}, 2019.

\bibitem[Grover et~al.()Grover, Tran, Shu, Poole, and Murphy]{groverprobing}
Grover, A., Tran, D., Shu, R., Poole, B., and Murphy, K.
\newblock Probing uncertainty estimates of neural processes.

\bibitem[Hakhamaneshi et~al.(2021)Hakhamaneshi, Abbeel, Stojanovic, and
  Grover]{hakhamaneshi2021jumbo}
Hakhamaneshi, K., Abbeel, P., Stojanovic, V., and Grover, A.
\newblock Jumbo: Scalable multi-task bayesian optimization using offline data.
\newblock \emph{arXiv preprint arXiv:2106.00942}, 2021.

\bibitem[Holderrieth et~al.(2021)Holderrieth, Hutchinson, and
  Teh]{holderrieth2021equivariant}
Holderrieth, P., Hutchinson, M.~J., and Teh, Y.~W.
\newblock Equivariant learning of stochastic fields: Gaussian processes and
  steerable conditional neural processes.
\newblock In \emph{International Conference on Machine Learning}, pp.\
  4297--4307. PMLR, 2021.

\bibitem[Kim et~al.(2019)Kim, Mnih, Schwarz, Garnelo, Eslami, Rosenbaum,
  Vinyals, and Teh]{kim2019attentive}
Kim, H., Mnih, A., Schwarz, J., Garnelo, M., Eslami, A., Rosenbaum, D.,
  Vinyals, O., and Teh, Y.~W.
\newblock Attentive neural processes.
\newblock \emph{arXiv preprint arXiv:1901.05761}, 2019.

\bibitem[Kim(2020)]{KimJ2020benchmark}
Kim, J.
\newblock Benchmark functions for bayesian optimization.
\newblock \url{https://github.com/jungtaekkim/bayeso-benchmarks}, 2020.

\bibitem[Kim \& Choi(2017)Kim and Choi]{KimJ2017bayeso}
Kim, J. and Choi, S.
\newblock {BayesO}: A {Bayesian} optimization framework in {Python}.
\newblock \url{https://bayeso.org}, 2017.

\bibitem[Kingma \& Welling(2013)Kingma and Welling]{kingma2013auto}
Kingma, D.~P. and Welling, M.
\newblock Auto-encoding variational bayes.
\newblock \emph{arXiv preprint arXiv:1312.6114}, 2013.

\bibitem[Kossen et~al.(2021)Kossen, Band, Lyle, Gomez, Rainforth, and
  Gal]{kossen2021self}
Kossen, J., Band, N., Lyle, C., Gomez, A.~N., Rainforth, T., and Gal, Y.
\newblock Self-attention between datapoints: Going beyond individual
  input-output pairs in deep learning.
\newblock \emph{arXiv preprint arXiv:2106.02584}, 2021.

\bibitem[Le et~al.(2018)Le, Kim, Garnelo, Rosenbaum, Schwarz, and
  Teh]{le2018empirical}
Le, T.~A., Kim, H., Garnelo, M., Rosenbaum, D., Schwarz, J., and Teh, Y.~W.
\newblock Empirical evaluation of neural process objectives.
\newblock In \emph{NeurIPS workshop on Bayesian Deep Learning}, 2018.

\bibitem[Lee et~al.(2019)Lee, Lee, Kim, Kosiorek, Choi, and Teh]{lee2019set}
Lee, J., Lee, Y., Kim, J., Kosiorek, A., Choi, S., and Teh, Y.~W.
\newblock Set transformer: A framework for attention-based
  permutation-invariant neural networks.
\newblock In \emph{International Conference on Machine Learning}, pp.\
  3744--3753. PMLR, 2019.

\bibitem[Lee et~al.(2020)Lee, Lee, Kim, Yang, Hwang, and
  Teh]{lee2020bootstrapping}
Lee, J., Lee, Y., Kim, J., Yang, E., Hwang, S.~J., and Teh, Y.~W.
\newblock Bootstrapping neural processes.
\newblock \emph{arXiv preprint arXiv:2008.02956}, 2020.

\bibitem[Lin et~al.(2020)Lin, Winata, Xu, Liu, and Fung]{lin2020variational}
Lin, Z., Winata, G.~I., Xu, P., Liu, Z., and Fung, P.
\newblock Variational transformers for diverse response generation.
\newblock \emph{arXiv preprint arXiv:2003.12738}, 2020.

\bibitem[Liu et~al.(2018)Liu, Luo, Wang, and Tang]{liu2018large}
Liu, Z., Luo, P., Wang, X., and Tang, X.
\newblock Large-scale celebfaces attributes (celeba) dataset.
\newblock \emph{Retrieved August}, 15\penalty0 (2018):\penalty0 11, 2018.

\bibitem[Lu et~al.(2021)Lu, Grover, Abbeel, and Mordatch]{lu2021pretrained}
Lu, K., Grover, A., Abbeel, P., and Mordatch, I.
\newblock Pretrained transformers as universal computation engines.
\newblock \emph{arXiv preprint arXiv:2103.05247}, 2021.

\bibitem[Mathieu et~al.(2021)Mathieu, Foster, and Teh]{mathieu2021contrastive}
Mathieu, E., Foster, A., and Teh, Y.~W.
\newblock On contrastive representations of stochastic processes.
\newblock \emph{arXiv preprint arXiv:2106.10052}, 2021.

\bibitem[Mockus et~al.(1978)Mockus, Tiesis, and
  Zilinskas]{mockus1978application}
Mockus, J., Tiesis, V., and Zilinskas, A.
\newblock The application of bayesian methods for seeking the extremum.
\newblock \emph{Towards global optimization}, 2\penalty0 (117-129):\penalty0 2,
  1978.

\bibitem[M{\"u}ller et~al.()M{\"u}ller, Hollmann, Arango, Grabocka, and
  Hutter]{mullertransformers}
M{\"u}ller, S., Hollmann, N., Arango, S.~P., Grabocka, J., and Hutter, F.
\newblock Transformers can do bayesian inference.
\newblock In \emph{International Conference on Learning Representations}.

\bibitem[Murphy et~al.(2018)Murphy, Srinivasan, Rao, and
  Ribeiro]{murphy2018janossy}
Murphy, R.~L., Srinivasan, B., Rao, V., and Ribeiro, B.
\newblock Janossy pooling: Learning deep permutation-invariant functions for
  variable-size inputs.
\newblock \emph{arXiv preprint arXiv:1811.01900}, 2018.

\bibitem[Radford et~al.(2018)Radford, Narasimhan, Salimans, and
  Sutskever]{radford2018improving}
Radford, A., Narasimhan, K., Salimans, T., and Sutskever, I.
\newblock Improving language understanding with unsupervised learning.
\newblock 2018.

\bibitem[Radford et~al.(2019)Radford, Wu, Child, Luan, Amodei, Sutskever,
  et~al.]{radford2019language}
Radford, A., Wu, J., Child, R., Luan, D., Amodei, D., Sutskever, I., et~al.
\newblock Language models are unsupervised multitask learners.
\newblock \emph{OpenAI blog}, 1\penalty0 (8):\penalty0 9, 2019.

\bibitem[Radford et~al.(2021)Radford, Kim, Hallacy, Ramesh, Goh, Agarwal,
  Sastry, Askell, Mishkin, Clark, et~al.]{radford2021learning}
Radford, A., Kim, J.~W., Hallacy, C., Ramesh, A., Goh, G., Agarwal, S., Sastry,
  G., Askell, A., Mishkin, P., Clark, J., et~al.
\newblock Learning transferable visual models from natural language
  supervision.
\newblock \emph{arXiv preprint arXiv:2103.00020}, 2021.

\bibitem[Riquelme et~al.(2018)Riquelme, Tucker, and Snoek]{riquelme2018deep}
Riquelme, C., Tucker, G., and Snoek, J.
\newblock Deep bayesian bandits showdown: An empirical comparison of bayesian
  deep networks for thompson sampling.
\newblock \emph{arXiv preprint arXiv:1802.09127}, 2018.

\bibitem[Schmidhuber(1987)]{schmidhuber1987evolutionary}
Schmidhuber, J.
\newblock \emph{Evolutionary principles in self-referential learning, or on
  learning how to learn: the meta-meta-... hook}.
\newblock PhD thesis, Technische Universit{\"a}t M{\"u}nchen, 1987.

\bibitem[Schonlau et~al.(1998)Schonlau, Welch, and Jones]{schonlau1998global}
Schonlau, M., Welch, W.~J., and Jones, D.~R.
\newblock Global versus local search in constrained optimization of computer
  models.
\newblock \emph{Lecture Notes-Monograph Series}, pp.\  11--25, 1998.

\bibitem[Shahriari et~al.(2015)Shahriari, Swersky, Wang, Adams, and
  De~Freitas]{shahriari2015taking}
Shahriari, B., Swersky, K., Wang, Z., Adams, R.~P., and De~Freitas, N.
\newblock Taking the human out of the loop: A review of bayesian optimization.
\newblock \emph{Proceedings of the IEEE}, 104\penalty0 (1):\penalty0 148--175,
  2015.

\bibitem[Singh et~al.(2019)Singh, Yoon, Son, and Ahn]{singh2019sequential}
Singh, G., Yoon, J., Son, Y., and Ahn, S.
\newblock Sequential neural processes.
\newblock \emph{arXiv preprint arXiv:1906.10264}, 2019.

\bibitem[Vanschoren(2018)]{vanschoren2018meta}
Vanschoren, J.
\newblock Meta-learning: A survey.
\newblock \emph{arXiv preprint arXiv:1810.03548}, 2018.

\bibitem[Vaswani et~al.(2017)Vaswani, Shazeer, Parmar, Uszkoreit, Jones, Gomez,
  Kaiser, and Polosukhin]{vaswani2017attention}
Vaswani, A., Shazeer, N., Parmar, N., Uszkoreit, J., Jones, L., Gomez, A.~N.,
  Kaiser, {\L}., and Polosukhin, I.
\newblock Attention is all you need.
\newblock In \emph{Advances in neural information processing systems}, pp.\
  5998--6008, 2017.

\bibitem[Xu et~al.(2020)Xu, Ton, Kim, Kosiorek, and Teh]{xu2020metafun}
Xu, J., Ton, J.-F., Kim, H., Kosiorek, A., and Teh, Y.~W.
\newblock Metafun: Meta-learning with iterative functional updates.
\newblock In \emph{International Conference on Machine Learning}, pp.\
  10617--10627. PMLR, 2020.

\bibitem[Zheng et~al.(2022)Zheng, Zhang, and Grover]{onlinedt}
Zheng, Q., Zhang, A., and Grover, A.
\newblock Online decision transformer.
\newblock In \emph{ICML}, 2022.

\end{thebibliography}
\bibliographystyle{icml2022}

\newpage
\appendix
\onecolumn
\icmltitle{Supplementary Materials to Transformer Neural Processes}
\section{Implementation details}
\subsection{Non-Diagonal Transformer Neural Processes} \label{sec:nd_details}
\begin{figure}[h]
    \centering
    \includegraphics[width=0.25\linewidth]{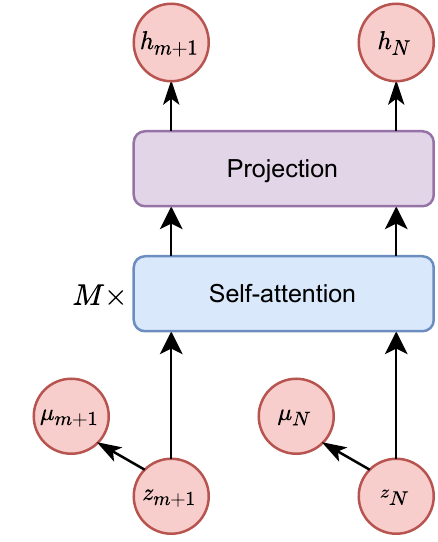}
    \caption{The decoder architecture of TNP-ND.}
    \label{fig:TNP-ND}
\end{figure}
Figure \ref{fig:TNP-ND} depicts the decoder architecture of TNP-NP presented in section \ref{sec:tnp_nd}, in which $\{z_{i}\}_{i=m+1}^N$ are the output of the last transformer decoder. Given $\{h_i\}_{i=m+1}^N$, the lower triangular matrix of the Multivariate Normal distribution is computed as:
\begin{equation}
    L = \text{lower}(H H^\top), \ H \in \mathbb{R}^{n \times p},
\end{equation}
in which $\text{lower}(L)$ removes the upper triangular parts of $L$, and $n=N-m$ is the number of the target points.

\subsection{Contextual bandits}
There is a crucial problem mismatch between training and evaluation. In training, we are given the reward values for all the arms, while during test time, we can only observe the reward value for the arm that we actually select, and the reward for all other arms will be drawn from a standard Gaussian distribution. To alleviate this problem, during training of TNPs, we randomly drop reward values for some arms of the context points and make the model to regress the ground-truth of those arms. This closes the gap between training and testing. We note that we also tried using this trick for other NP variants but it did not improve the performance.

\subsection{Hyperparameters}
In this section we present the hyperparameters that we used to train TNPs in the experiments. For the baselines, we use the hyperparameters reported in \citet{lee2020bootstrapping}.
\paragraph{1-D regression}
\begin{itemize}
    \item Model dimension: $64$
    \item Number of embeddings layers: $4$
    \item Feed forward dimension: $128$
    \item Number of attention heads: $4$
    \item Number of transformer layers: $6$
    \item (TNP-ND) Number of self-attention layers on $\{z_i\}_{i=m+1}^N$: $2$
    \item (TNP-ND) Projection dimension: $20$
    \item (TNP-ND) Number of projection layers: $4$
    \item Dropout: $0.0$
    \item Number of training steps: $100000$
    \item Learning rate: $5e^{-4}$ with Cosine annealing scheduler
\end{itemize}

\paragraph{Image completion}
\begin{itemize}
    \item Model dimension: $64$
    \item Number of embeddings layers: $4$
    \item Feed forward dimension: $128$
    \item Number of attention heads: $4$
    \item Number of transformer layers: $6$
    \item (TNP-ND) Number of self-attention layers on $\{z_i\}_{i=m+1}^N$: $2$
    \item (TNP-ND) Projection dimension: $20$
    \item (TNP-ND) Number of projection layers: $4$
    \item Dropout: $0.0$
    \item Batch size: $100$
    \item Number of training epochs: $200$
    \item Learning rate: $5e^{-4}$ with Cosine annealing scheduler
\end{itemize}

\paragraph{Contextual bandits}
\begin{itemize}
    \item Model dimension: $16$
    \item Number of embeddings layers: $3$
    \item Feed forward dimension: $64$
    \item Number of attention heads: $1$
    \item Number of transformer layers: $4$
    \item (TNP-ND) Number of self-attention layers on $\{z_i\}_{i=m+1}^N$: $2$
    \item (TNP-ND) Projection dimension: $20$
    \item (TNP-ND) Number of projection layers: $4$
    \item Dropout: $0.0$
    \item Number of training epochs: $100000$
    \item Learning rate: $5e^{-4}$ with Cosine annealing scheduler
    \item Drop rate of reward values during training: $0.5$
\end{itemize}

\paragraph{Bayesian optimization}
\begin{itemize}
    \item Model dimension: $64$
    \item Number of embeddings layers: $4$
    \item Feed forward dimension: $128$
    \item Number of attention heads: $8$
    \item Number of transformer layers: $6$
    \item (TNP-ND) Number of self-attention layers on $\{z_i\}_{i=m+1}^N$: $2$
    \item (TNP-ND) Projection dimension: $20$
    \item (TNP-ND) Number of projection layers: $4$
    \item Dropout: $0.0$
    \item Number of training steps: $100000$
    \item Learning rate: $5e^{-4}$ with Cosine annealing scheduler
\end{itemize}

\section{Additional results}
\subsection{1-D regression with additional metrics} \label{sec:1d_additional}
We compare TNPs and the baselines on various metrics, which include root-mean-square error, calibration error, and log-likelihood. The metrics are all computed on the predictions of the target points by using the Uncertainty Toolbox \citep{chung2021uncertainty}.

Table \ref{tab:1d_more_metrics} shows that three variants of TNPs outperform the baselines on all metrics in $2/3$ kernels. There is an interesting pattern among the NP-based methods: while using attention nearly always leads to better accuracy (CANPs, ANPs, BANPs versus CNPs, NPs, BNPs), it often results in poorer calibration. TNPs have the best of both worlds, since they are competitively accurate and calibrated in most tasks. The three variants achieve similar accuracy and degree of calibration, while TNP-A is the best variant with respect to log-likelihood metric.

\begin{table}[ht]
\centering
\caption{Comparison of TNPs with the baselines on various GP kernels and evaluation metrics. We train $5$ instances with different seeds for each method and report the mean and std.}
\label{tab:1d_more_metrics}
\scalebox{1.0}{
\begin{tabular}{ccccc}
\toprule
Metric                          & Method       & RBF            & Matérn 5/2     & Periodic   \\
\midrule
\multirow{9}{*}{RMSE}           & CNP          & $0.278 \pm 0.003$ & $0.310 \pm 0.003$ & $0.652 \pm 0.001$      \\  
                                & CANP         & $0.193 \pm 0.000$ & $0.228 \pm 0.000$ & $0.699 \pm 0.002$    \\ 
                                & NP           & $0.282 \pm 0.003$ & $0.315 \pm 0.003$ & $0.650 \pm 0.002$     \\  
                                & ANP          & $0.193 \pm 0.001$ & $0.230 \pm 0.000$ & $0.703 \pm 0.002$     \\  
                                & BNP          & $0.269 \pm 0.003$ & $0.301 \pm 0.003$ & $\boldsymbol{0.649 \pm 0.002}$  \\  
                                & BANP         & $0.192 \pm 0.001$ & $0.228 \pm 0.001$ & $0.701 \pm 0.008$   \\  
                                & TNP-D         & $\boldsymbol{0.177 \pm 0.001}$  & $\boldsymbol{0.222 \pm 0.000}$  & $0.664 \pm 0.014$  \\
                                & TNP-A          & $0.178 \pm 0.000$  & $\boldsymbol{0.222 \pm 0.000}$  & $0.660 \pm 0.002$   \\
                                & TNP-ND & $0.180 \pm 0.001$  & $0.223 \pm 0.000$  & $0.670 \pm 0.009$  \\
\midrule
\multirow{9}{*}{CE}             & CNP          & $0.078 \pm 0.002$ & $0.051 \pm 0.000$ & $0.143 \pm 0.002$   \\  
                                & CANP         & $0.232 \pm 0.001$ & $0.165 \pm 0.000$ & $0.255 \pm 0.004$    \\  
                                & NP           & $0.093 \pm 0.002$ & $0.056 \pm 0.001$ & $0.130 \pm 0.007$   \\  
                                & ANP          & $0.235 \pm 0.001$ & $0.169 \pm 0.001$ & $0.265 \pm 0.002$    \\  
                                & BNP          & $0.093 \pm 0.003$ & $0.054 \pm 0.002$ & $\boldsymbol{0.115 \pm 0.004}$  \\  
                                & BANP         & $0.236 \pm 0.002$ & $0.171 \pm 0.002$ & $0.217 \pm 0.005$    \\  
                                & TNP-D         & $\boldsymbol{0.043 \pm 0.000}$  & $0.045 \pm 0.000$  & $0.129 \pm 0.012$ \\
                                & TNP-A          & $0.045 \pm 0.000$  & $\boldsymbol{0.044 \pm 0.000}$  & $0.119 \pm 0.008$  \\
                                & TNP-ND & $0.048 \pm 0.001$  & $0.050 \pm 0.001$  & $0.155 \pm 0.009$  \\
\midrule
\multirow{9}{*}{Log-Likelihood} & CNP          & $0.26 \pm 0.02$ & $0.04 \pm 0.02$ & $-1.40 \pm 0.02$  \\  
                                & CANP         & $0.79 \pm 0.00$ & $0.62 \pm 0.00$ & $-7.61 \pm 0.16$   \\  
                                & NP           & $0.27 \pm 0.01$ & $0.07 \pm 0.01$ & $-1.15 \pm 0.04$   \\  
                                & ANP          & $0.81 \pm 0.00$ & $0.63 \pm 0.00$ & $-5.02 \pm 0.21$   \\  
                                & BNP          & $0.38 \pm 0.02$ & $0.18 \pm 0.02$ & $\boldsymbol{-0.96 \pm 0.02}$ \\  
                                & BANP         & $0.82 \pm 0.01$ & $0.66 \pm 0.00$ & $-3.09 \pm 0.14$ \\  
                                & TNP-D         & $1.39 \pm 0.00$  & $0.95 \pm 0.01$  & $-3.53 \pm 0.37$    \\
                                & TNP-A          & $\boldsymbol{1.63 \pm 0.00}$  & $\boldsymbol{1.21 \pm 0.00}$  & $-2.26 \pm 0.17$  \\
                                & TNP-ND & $1.46 \pm 0.00$  & $1.02 \pm 0.00$  & $-4.13 \pm 0.33$   \\
\bottomrule
\end{tabular}
}
\end{table}

\newpage
\subsection{Image completion} \label{image_add}
\paragraph{Additional metrics} We compare TNPs and the baselines on the metrics used in Section \ref{sec:1d_additional}, except for the Calibration error which took too much time to compute.
Tables \ref{tab:celeba_add} and \ref{tab:emnist_add} show the results for CelebA and EMNIST, respectively. It is clear that three variants of TNPs outperform the baselines on both datasets on all 3 evaluation metrics. While three variants are similar in terms of accuracy and sharpness, TNP-A is the best variant with respect to log-likelihood.
\begin{table}[h!]
\centering
\caption{Comparison of TNPs vs the baselines on CelebA dataset with various evaluation metrics. We train $5$ instances with different seeds for each method and report the mean and std.}
\label{tab:celeba_add}
\scalebox{1.0}{
\begin{tabular}{cccc}
\toprule
Method       & RMSE          & Log-likelihood \\
\midrule
CNP          & $0.137 \pm 0.000$ & $2.148 \pm 0.008$ \\
CANP         & $0.116 \pm 0.001$ & $2.657 \pm 0.010$ \\
NP           & $0.138 \pm 0.001$ & $2.480 \pm 0.018$ \\
ANP          & $0.119 \pm 0.000$ & $2.904 \pm 0.003$ \\
BNP          & $0.134 \pm 0.000$ & $2.764 \pm 0.006$ \\
BANP         & $0.119 \pm 0.000$ & $3.087 \pm 0.004$ \\
TNP-D        & $0.112 \pm 0.000$ & $3.891 \pm 0.006$ \\ 
TNP-A        & $0.115 \pm 0.000$ & $\boldsymbol{5.818 \pm 0.011}$  \\
TNP-ND       & $\boldsymbol{0.111 \pm 0.000}$ & $5.477 \pm 0.016$  \\
\bottomrule
\end{tabular}
}
\end{table}

\begin{table}[h!]
\centering
\caption{Comparison of TNPs vs the baselines on EMNIST dataset with various evaluation metrics. We train $5$ instances with different seeds for each method and report the mean and std. We evaluate on both seen and unseen classes.}
\label{tab:emnist_add}
\begin{tabular}{@{}cccc@{}}
\toprule
Setting                                                                           & Method       & RMSE           & Log-likelihood \\
\midrule
\multirow{9}{*}{\begin{tabular}[c]{@{}c@{}}Seen classes\\ (0-9)\end{tabular}}     & CNP          & $0.184 \pm 0.001$ & $0.733 \pm 0.004$ \\
                                                                                  & CANP         & $0.140 \pm 0.002$ & $0.935 \pm 0.008$ \\
                                                                                  & NP           & $0.185 \pm 0.003$ & $0.794 \pm 0.010$ \\
                                                                                  & ANP          & $0.142 \pm 0.000$ & $0.984 \pm 0.003$ \\
                                                                                  & BNP          & $0.178 \pm 0.002$ & $0.876 \pm 0.007$ \\
                                                                                  & BANP         & $0.142 \pm 0.001$ & $1.008 \pm 0.003$ \\
                                                                                  & TNP-D & $0.119 \pm 0.002$  & $1.461 \pm 0.010$  \\
                                                                                  & TNP-A  & $0.122 \pm 0.001$  & $\boldsymbol{1.537 \pm 0.005}$  \\
                                                                                  & TNP-ND & $\boldsymbol{0.116 \pm 0.000}$ & $1.497 \pm 0.002$  \\
\midrule
\multirow{9}{*}{\begin{tabular}[c]{@{}c@{}}Unseen classes\\ (10-46)\end{tabular}} & CNP          & $0.225 \pm 0.002$ & $0.491 \pm 0.010$ \\
                                                                                  & CANP         & $0.163 \pm 0.002$ & $0.823 \pm 0.010$ \\
                                                                                  & NP           & $0.228 \pm 0.003$ & $0.591 \pm 0.009$ \\
                                                                                  & ANP          & $0.166 \pm 0.001$ & $0.887 \pm 0.004$ \\
                                                                                  & BNP          & $0.219 \pm 0.003$ & $0.728 \pm 0.008$ \\
                                                                                  & BANP         & $0.161 \pm 0.001$ & $0.943 \pm 0.003$ \\
                                                                                  & TNP-D & $\boldsymbol{0.139 \pm 0.001}$  & $1.308 \pm 0.003$ \\
                                                                                  & TNP-A & $0.142 \pm 0.001$  & $\boldsymbol{1.413 \pm 0.005}$  \\
                                                                                  & TNP-ND & $0.140 \pm 0.001$ & $1.314 \pm 0.004$  \\
\bottomrule
\end{tabular}
\end{table}

\paragraph{Evaluation on full-image completion}  
In Tables~\ref{tab:img_benchmark},~\ref{tab:celeba_add}, and~\ref{tab:emnist_add}, we computed the metrics on a target set, which is a subset of the entire image. We additionally report the performance of TNP-D and BANP (the strongest baseline) when the target is the entire image in Table~\ref{tab:emnist_entire_image}, which shows that TNPs achieve a similarly better performance.
\begin{table}[h!]
\centering
\caption{Comparison of TNP-D and BANP on EMNIST unseen classes. We use 100 context points, and the target is the entire image.}
\label{tab:emnist_entire_image}
\begin{tabular}{@{}lcc@{}}
\toprule
Method & RMSE              & Log-likelihood     \\ \midrule
BANP  & $0.144 \pm 0.001$ & $1.005 \pm 0.003$ \\
TNP-D & $\mathbf{0.117 \pm 0.001}$ & $\mathbf{1.466 \pm 0.005}$  \\ \bottomrule
\end{tabular}
\end{table}

\paragraph{Comparison with Conv(C)NP}
ConvCNP and ConvNP~\cite{gordon2019convolutional,foong2020meta} build translation equivariance into NPs, which should be helpful in specific domains such as images. However, table \ref{tab:image_conv} shows that even in image completion, TNPs still outperform Conv(C)NP by a large margin. For off-grid data, Conv(C)NP is only applicable when $x$ is one-dimensional, as they require a discretization step, limiting their application to high-dimensional BO and contextual bandits.
\begin{table}[h!]
\centering
\caption{TNP vs Conv(C)NP on EMNIST image completion. We train $5$ instances with different seeds for each method and report the mean and std.}
\label{tab:image_conv}
\begin{tabular}{@{}lcc@{}}
\toprule
Method  & Seen classes (0-9)     & Unseen classes (10-46) \\ \midrule
ConvCNP & $1.02 \pm 0.00$        & $0.92 \pm 0.00$        \\
ConvNP  & $1.15 \pm 0.00$        & $1.10 \pm 0.00$       \\
TNP-A   & $\mathbf{1.54 \pm 0.01}$ & $\mathbf{1.41 \pm 0.01}$ \\
TNP-D   & $1.46 \pm 0.01$        & $1.31 \pm 0.00$        \\
TNP-ND  & $1.50 \pm 0.00$        & $1.31 \pm 0.00$        \\ \bottomrule
\end{tabular}
\end{table}

\newpage
\paragraph{Qualitative results} We visualize the completion of randomly selected images produced by TNPs and the baselines for both CelebA and EMNIST datasets.

\begin{figure}[ht]
    \centering
    \includegraphics[width=0.9\textwidth]{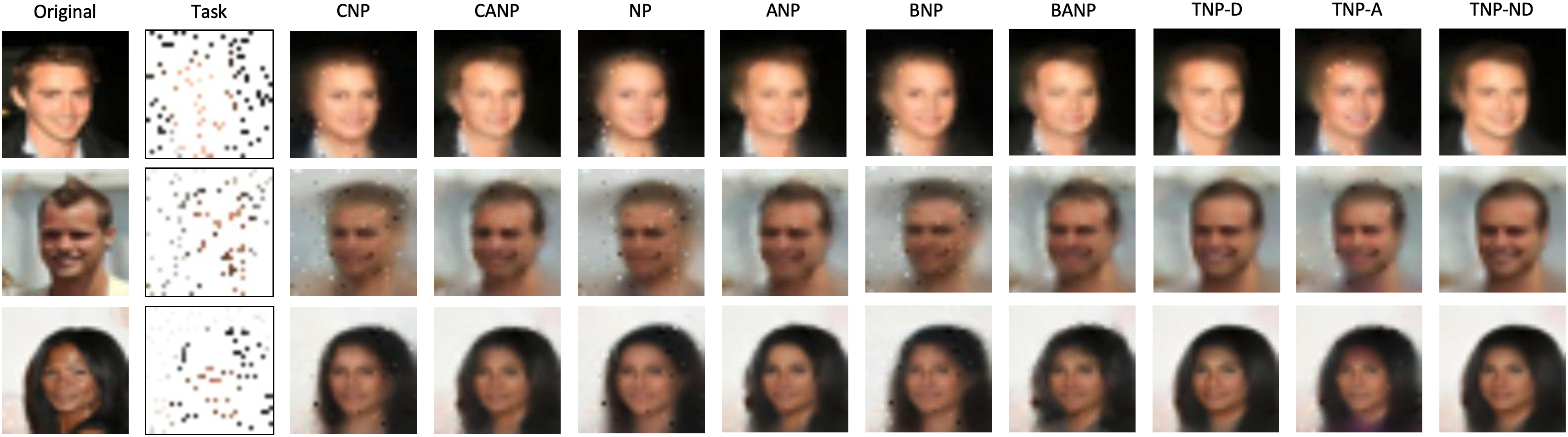}
    \caption{Completed images produced by TNPs and the baselines. The number of context points is 100.}
    \label{fig:celeba_complete}
\end{figure}

\begin{figure}[h!]
    \centering
    \includegraphics[width=0.9\textwidth]{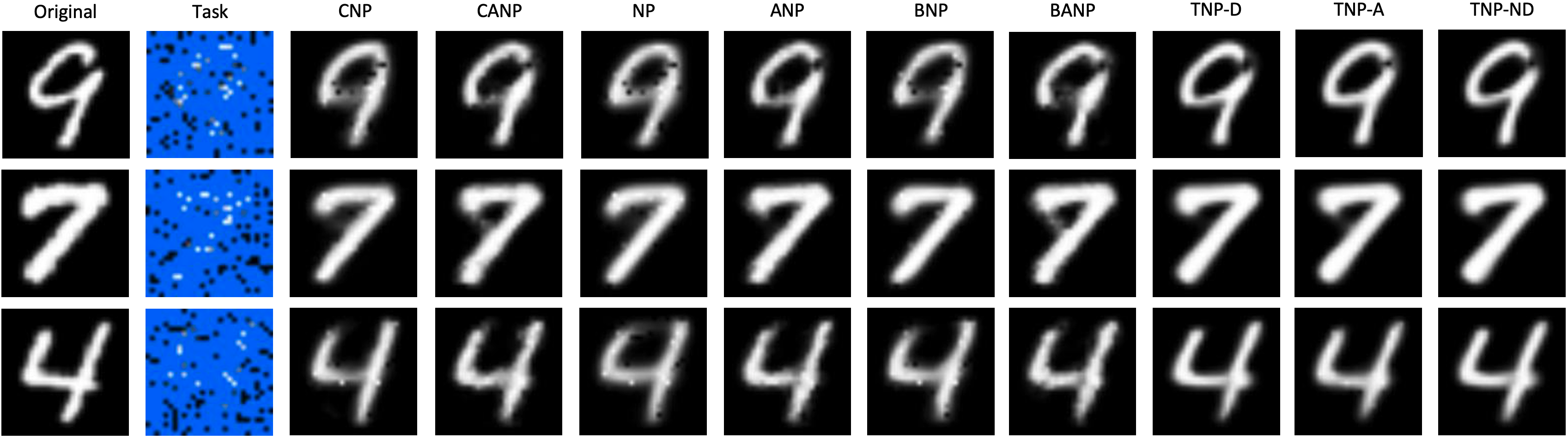}
    \caption{Image completion on seen classes by TNPs and the baselines. The number of context points is 100.}
    \label{fig:emnist_complete_1}
\end{figure}

\begin{figure}[h!]
    \centering
    \includegraphics[width=0.9\textwidth]{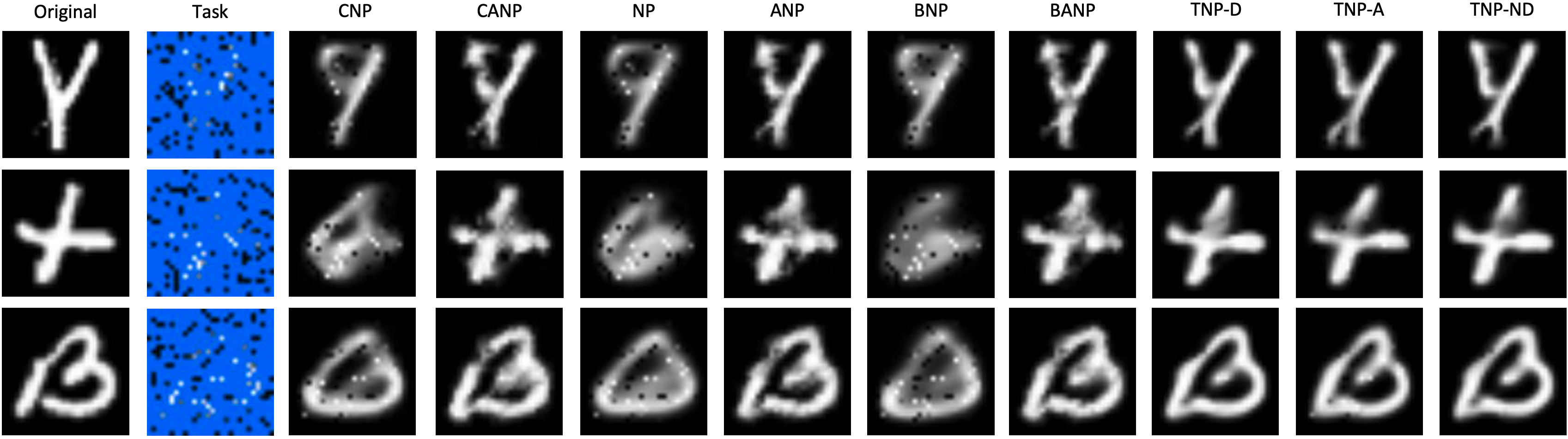}
    \caption{Image completion on unseen classes TNPs and the baselines. The number of context points is 100.}
    \label{fig:emnist_complete_2}
\end{figure}

Figures \ref{fig:celeba_complete}, \ref{fig:emnist_complete_1}, and \ref{fig:emnist_complete_2} show the visualizations. It is clear from the visualizations that the variants of TNPs produce more accurate images and with less artifacts. This is especially evident when we look at Figure \ref{fig:emnist_complete_2} which shows the completed images for unseen classes. TNPs are the only method that produced interpretable letters.


\newpage

\subsection{Contextual bandit} \label{sec:bandit_add}
\begin{table*}[h]
\centering
\caption{Comparison of TNPs with the baselines on simple regret on contextual bandit problems with different values of $\delta$. We run each model $50$ times for each value of $\delta$ and report the mean and standard deviation.}
\scalebox{0.9}{
\begin{tabular}{ccccccc}
\toprule
Method       & $\delta=0.7$   & $\delta=0.9$   & $\delta=0.95$  & $\delta=0.99$  & $\delta=0.995$  & $\delta=0.999$ \\ 
\midrule
Uniform      & $100.00 \pm 20.77$ & $100.00 \pm 34.60$ & $100.00 \pm 50.34$ & $100.00 \pm 96.59$ & $100.00 \pm 114.30$ & $100.00 \pm 120.11$ \\ 
CNP          & $4.06 \pm 4.15$    & $8.13 \pm 9.44$    & $8.06 \pm 9.46$    & $28.05 \pm 19.91$  & $39.50 \pm 27.30$   & $93.57 \pm 64.70$   \\ 
CANP         & $1.25 \pm 2.42$    & $2.97 \pm 5.39$    & $8.82 \pm 14.01$   & $41.01 \pm 69.41$  & $46.49 \pm 88.31$   & $34.62 \pm 115.25$  \\ 
NP           & $1.65 \pm 2.43$    & $2.95 \pm 4.07$    & $4.45 \pm 4.56$    & $18.64 \pm 1.87$   & $26.14 \pm 2.65$    & $62.13 \pm 6.21$    \\ 
ANP          & $1.58 \pm 2.22$    & $3.92 \pm 5.10$    & $5.38 \pm 5.66$    & $19.69 \pm 1.88$   & $27.44 \pm 2.63$    & $66.39 \pm 6.34$    \\ 
BNP          & $62.72 \pm 15.44$  & $57.22 \pm 25.55$  & $58.58 \pm 37.25$  & $63.02 \pm 75.67$  & $65.77 \pm 84.45$   & $78.07 \pm 94.31$   \\ 
BANP         & $3.32 \pm 3.12$    & $11.31 \pm 14$  & $34.19 \pm 28.13$  & $65.60 \pm 91.17$  & $59.33 \pm 101.53$  & $30.13 \pm 71.82$   \\ 
TNP-D & $\boldsymbol{0.67 \pm 3}$    & $1.41 \pm 2.86$    & $2.24 \pm 5.46$    & $3.13 \pm 1.21$    & $4.31 \pm 1.75$     & $9.16 \pm 3.86$     \\
TNP-A          & $1.17 \pm 1.73$ & $2.53 \pm 5.03$ & $2.47 \pm 6.83$ & $\boldsymbol{2.12 \pm 5.66}$ & $\boldsymbol{3.53 \pm 13.42}$ & $\boldsymbol{5.07 \pm 2.98}$ \\
TNP-ND & $1.40 \pm 2.33$ & $\boldsymbol{0.78 \pm 2.30}$ & $\boldsymbol{0.63 \pm 0.26}$ & $3.22 \pm 1.25$ & $4.86 \pm 1.88$ & $11.08 \pm 4.36$ \\
\bottomrule
\end{tabular}
}
\end{table*}

\section{TNPs vs other NP variants on sampling multiple functions}
An advantage of using latent variables is the ability to sample diverse functions from the same set of context points. However, we can also sample multiple functions using an autoregressive sequence model. In this section, we present how we obtain samples from TNP-A, and qualitatively measure the sample diversity of TNP-A and the baselines on 1-D regression and image completion tasks.

\subsection{Sampling procedure of TNP-A}
We sequentially sample the target points $y_{m+1:N}$ given the target inputs $x_{m+1:N}$ and the set of context points $\{x_i, y_i\}_{i=1}^m$. Specifically, for each step $i = 1$ to $i = N - m$, we:
\begin{enumerate}
    \item Obtain the predictive distribution $p(y_{m+i} \mid x_{1:m+i}, y_{1:m}, \hat{y}_{m+1:m+i-1})$, in which $\hat{y}_{m+1:m+i-1}$ are samples of the previous target points.
    \item Sample $\hat{y}_{m+i} \sim p(y_{m+i} \mid x_{1:m+i}, y_{1:m}, \hat{y}_{m+1:m+i-1})$.
    \item Repeat
\end{enumerate}

\subsection{Sampling results}
\begin{figure}[h!]
     \centering
     \begin{subfigure}[t]{0.78\textwidth}
         \centering
         \includegraphics[width=\textwidth]{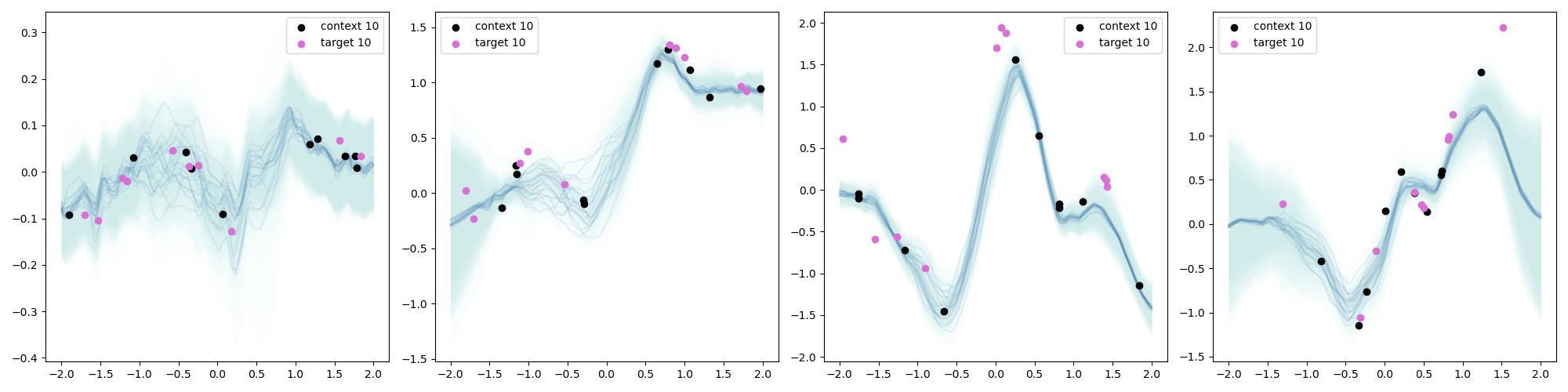}
         \caption{Sample functions produced by NPs.}
     \end{subfigure}
     \begin{subfigure}[t]{0.78\textwidth}
         \centering
         \includegraphics[width=\textwidth]{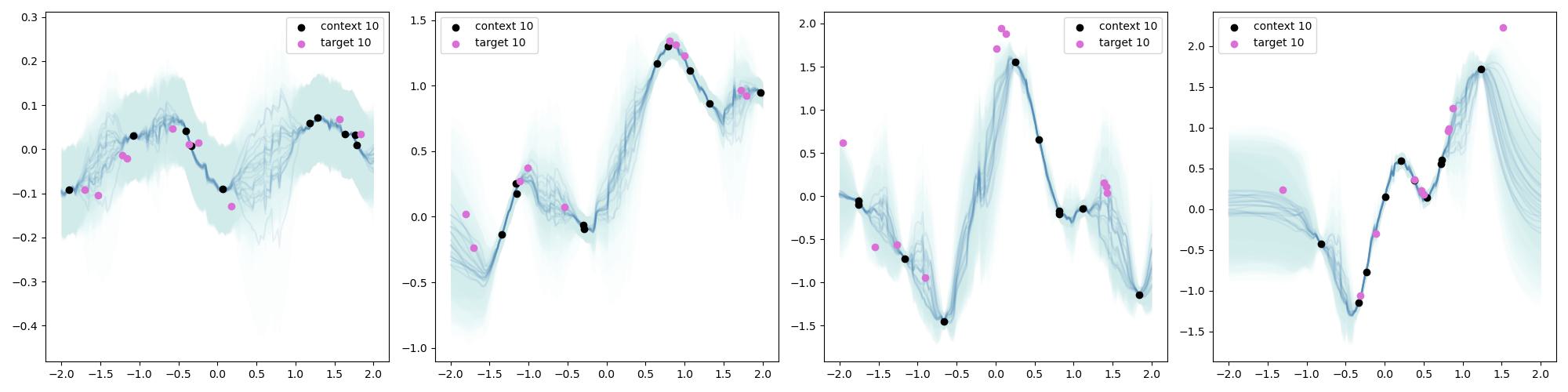}
         \caption{Samples functions produced by ANPs.}
     \end{subfigure}
     \begin{subfigure}[t]{0.78\textwidth}
         \centering
         \includegraphics[width=\textwidth]{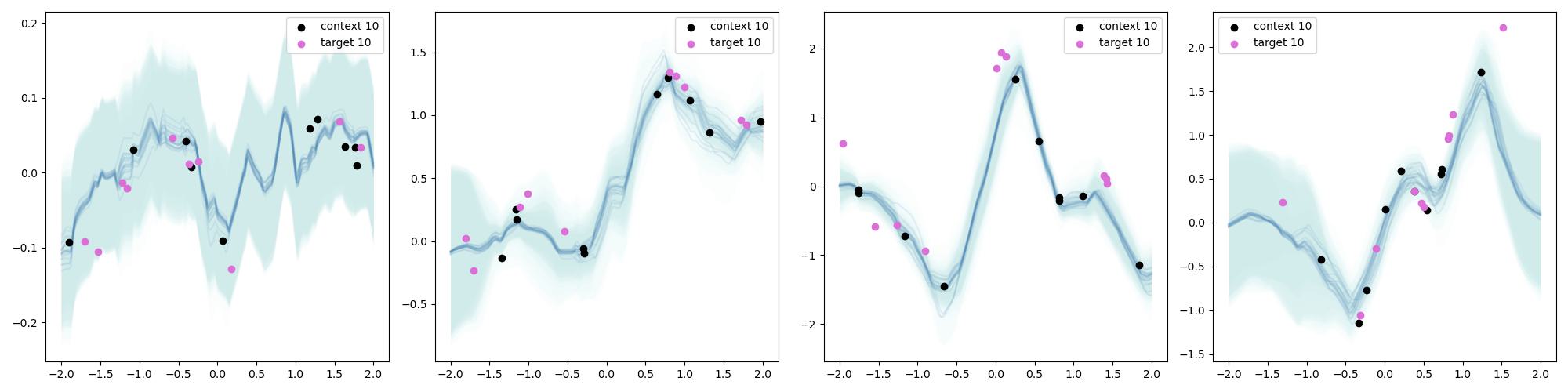}
         \caption{Samples functions produced by BNPs.}
     \end{subfigure}
     \begin{subfigure}[t]{0.78\textwidth}
         \centering
         \includegraphics[width=\textwidth]{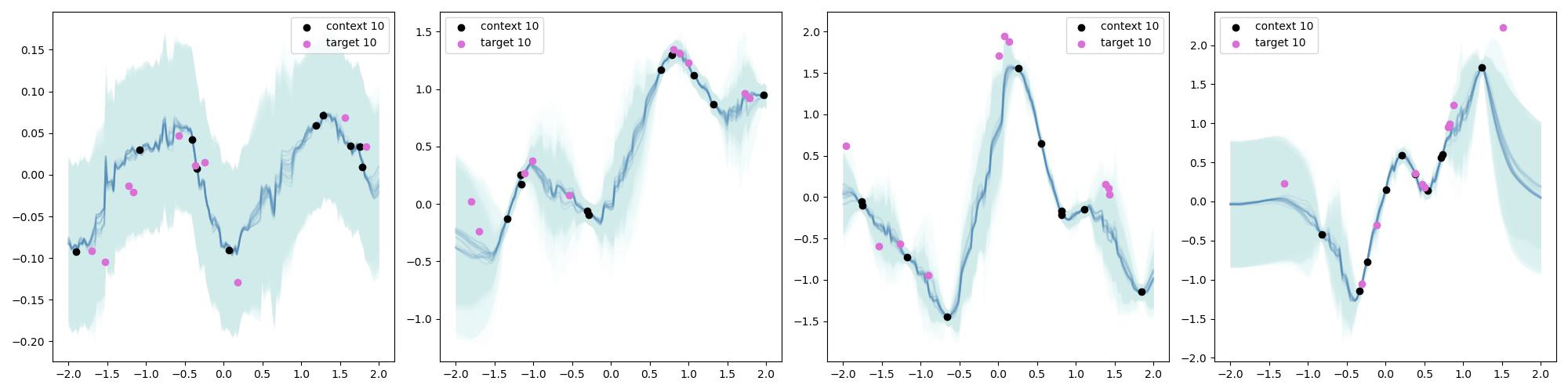}
         \caption{Samples functions produced by BANPs.}
     \end{subfigure}
     \begin{subfigure}[t]{0.78\textwidth}
         \centering
         \includegraphics[width=\textwidth]{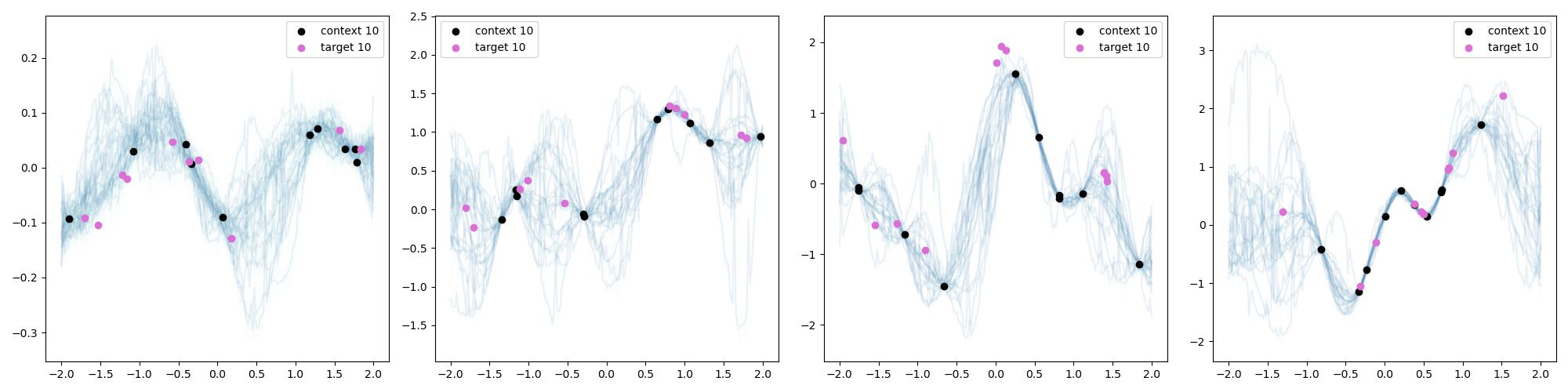}
         \caption{Samples functions produced by TNP-A.}
     \end{subfigure}
    \caption{Sample functions produced by TNPs and the baselines given $10$ context points. Data is generated from a GP with an RBF kernel. Each solid blue curve corresponds to one sample function, and the blue area around each curve represents the variance in the predictive distribution.}
    \label{fig:1d_samples_10}
\end{figure}

\begin{figure}[h!]
     \centering
     \begin{subfigure}[t]{0.8\textwidth}
         \centering
         \includegraphics[width=\textwidth]{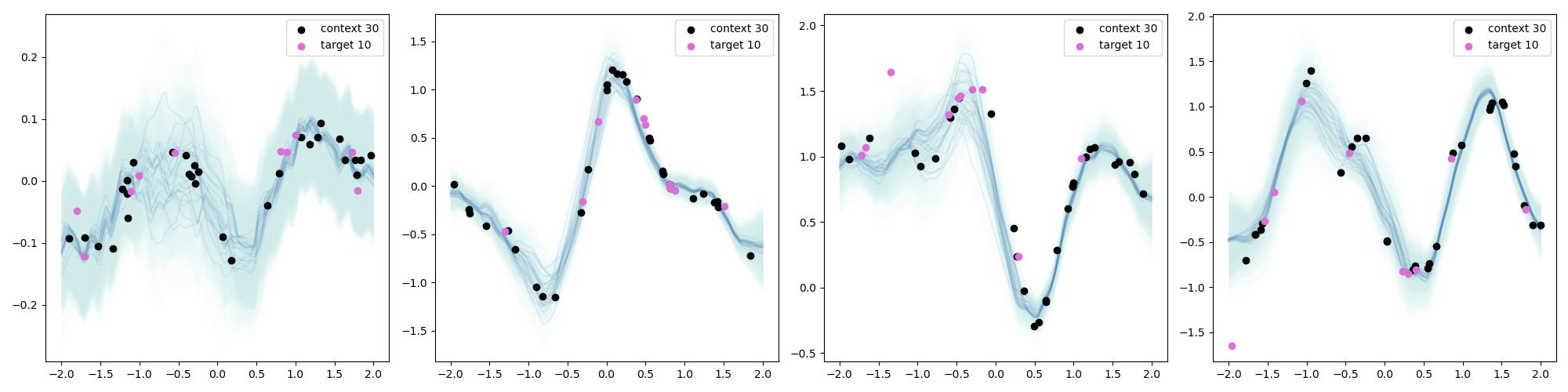}
         \caption{Sample functions produced by NPs.}
     \end{subfigure}
     \begin{subfigure}[t]{0.8\textwidth}
         \centering
         \includegraphics[width=\textwidth]{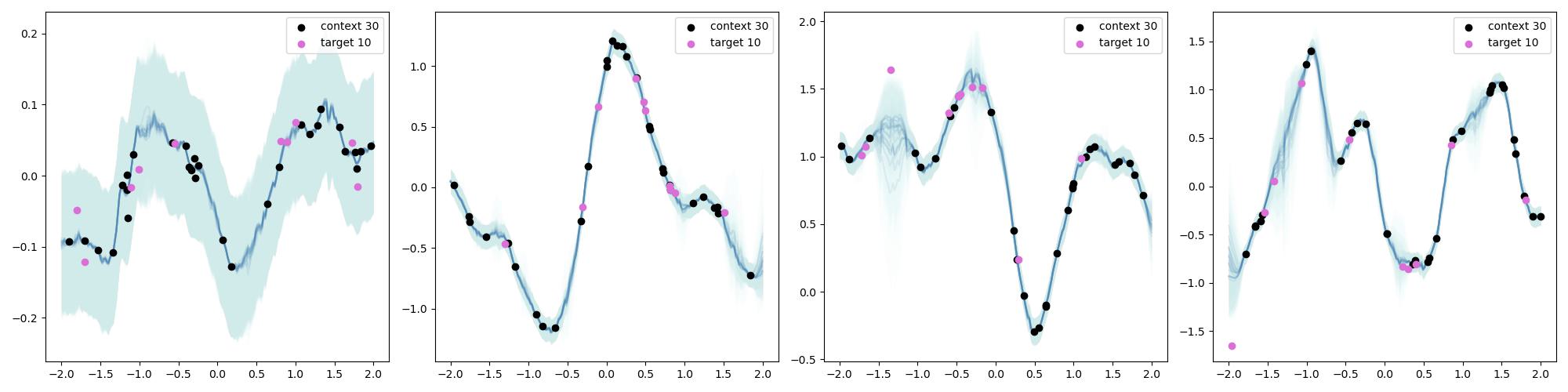}
         \caption{Samples functions produced by ANPs.}
     \end{subfigure}
     \begin{subfigure}[t]{0.8\textwidth}
         \centering
         \includegraphics[width=\textwidth]{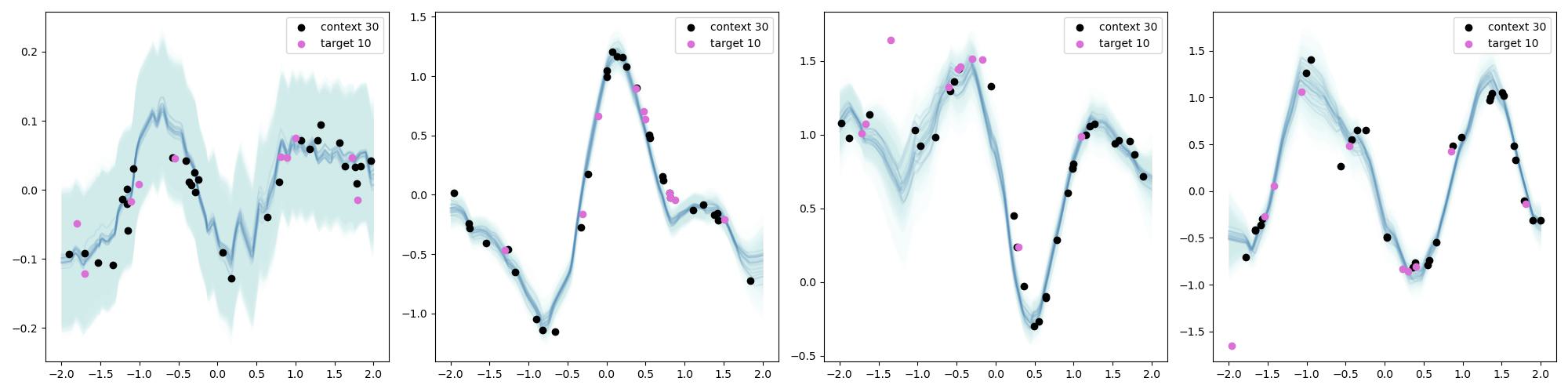}
         \caption{Samples functions produced by BNPs.}
     \end{subfigure}
     \begin{subfigure}[t]{0.8\textwidth}
         \centering
         \includegraphics[width=\textwidth]{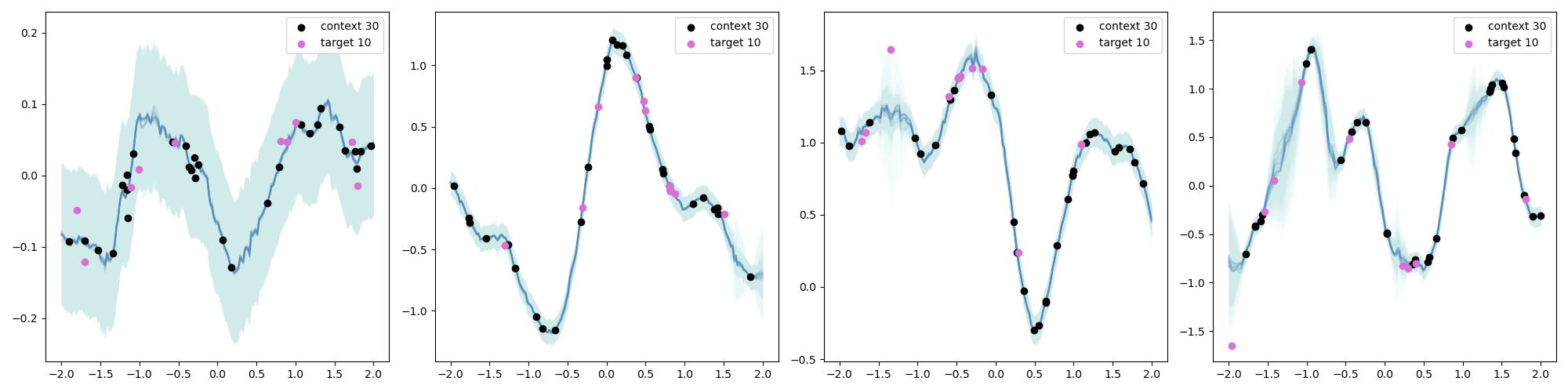}
         \caption{Samples functions produced by BANPs.}
     \end{subfigure}
     \begin{subfigure}[t]{0.8\textwidth}
         \centering
         \includegraphics[width=\textwidth]{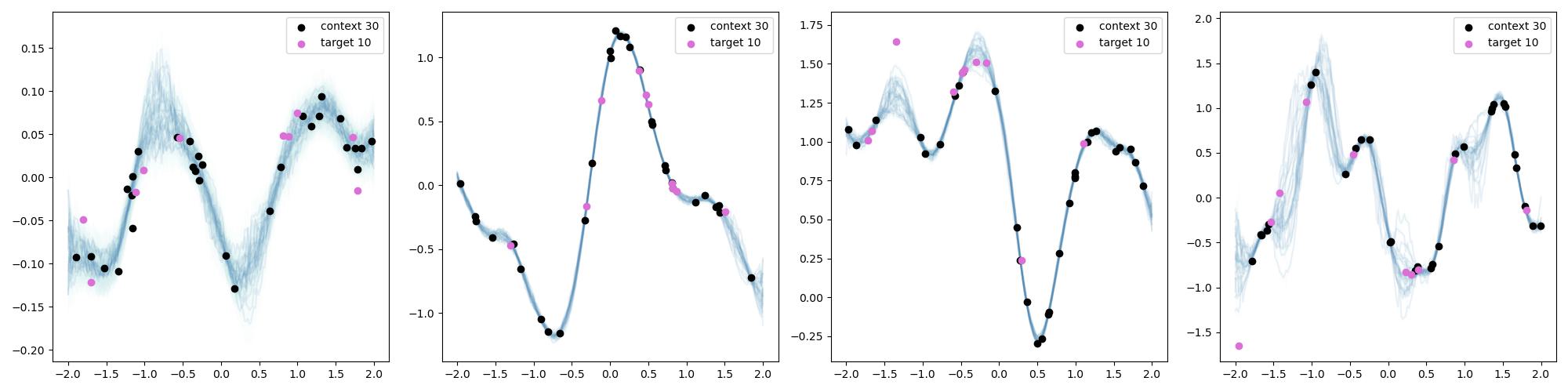}
         \caption{Samples functions produced by TNP-A.}
     \end{subfigure}
    \caption{Sample functions produced by TNPs and the baselines given $30$ context points. Data is generated from a GP with an RBF kernel. Each solid blue curve corresponds to one sample function, and the blue area around each curve represents the variance in the predictive distribution.}
    \label{fig:1d_samples_30}
\end{figure}

\paragraph{1-D regression} Figure \ref{fig:1d_samples_10} and \ref{fig:1d_samples_30} show sample functions produced by NP, ANP, BNP, BANP, and TNP-A given $10$ and $30$ context points, respectively. Each method is tested on the same set of $4$ underlying functions, which are sampled from a GP with an RBF kernel. In the figures, each solid blue curve is a sample function, and the blue area around the curve represents the variance of the predictive distribution over $y$. We can see that when the number of context points is $10$, all methods can produce different samples. However, TNP-A tends to have more diverse samples (more diverse solid curves), while the other NP methods use the predictive variance to account for their uncertainty, which result in large blue areas. When there are $30$ context points, all methods produce consistent functions.

\newpage

\newpage

\newpage

\begin{figure}[h!]
     \centering
     \begin{subfigure}[t]{0.4\textwidth}
         \centering
         \includegraphics[width=\textwidth]{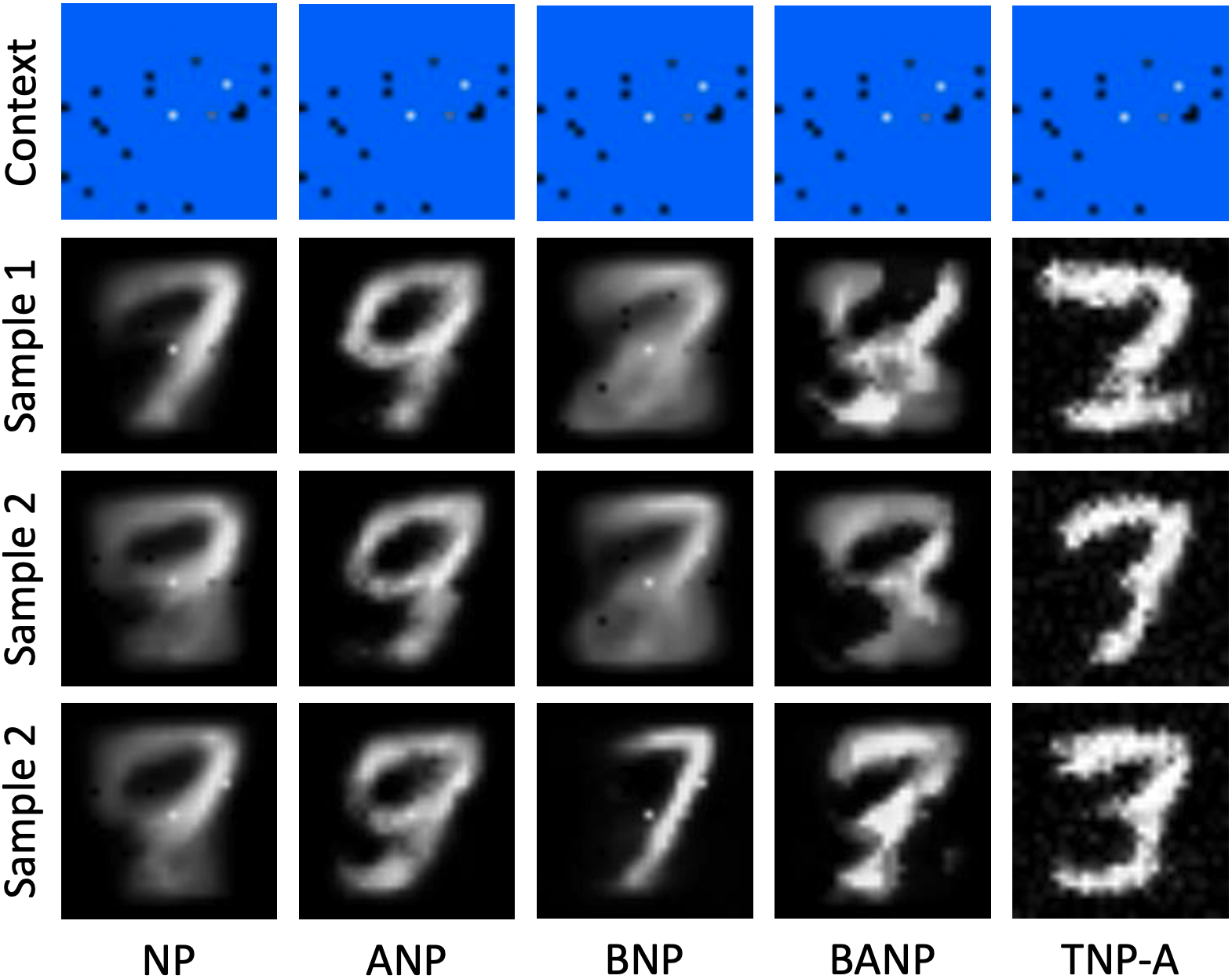}
         \caption{Samples produced by TNPs and the baselines given $20$ context points.}
     \end{subfigure}
     \hspace{0.7cm}
     \begin{subfigure}[t]{0.4\textwidth}
         \centering
         \includegraphics[width=\textwidth]{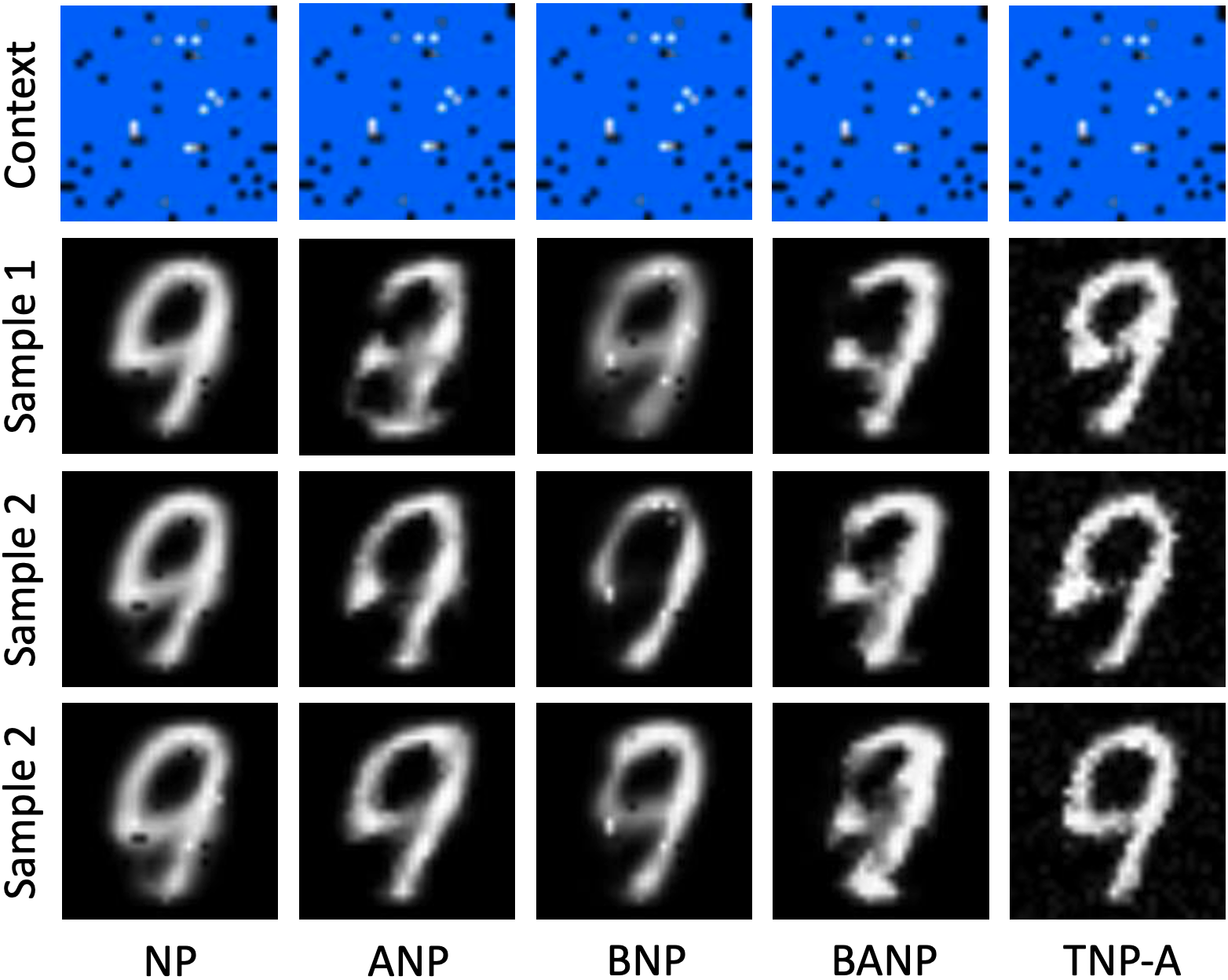}
         \caption{Samples produced by TNPs and the baselines given $50$ context points.}
     \end{subfigure}
    \caption{Sample images produced by TNPs and the baselines given the same set of context points. The original image is drawn randomly from the test set.}
    \label{fig:emnist_sample}
\end{figure}

\paragraph{Image completion} Figure \ref{fig:emnist_sample} show different completed images produced by NP, ANP, BNP, BANP, and TNP-A given $20$ and $50$ context points, respectively. When the number of context points is only $20$, there are multiple possible digits that can be generated from the same set of context points. TNP-A produces three samples that represent three different digits, while the diversity of other NP methods is not clear. When the number of context points is $50$, all methods produce consistent samples that represent the digit $9$.

\section{Architectural ablation analysis}
Before coming up with the final architectures of TNPs, we had experimented with other possible architectures. In this section, we present these alternative design choices, and their performances compared to TNPs.

\subsection{TNPs with alternative input representations}
There are other input representations to TNPs in addition to concatenating $x$ and $y$. We present these alternatives in this section and show how they perform compared to TNPs. We show ablation for TNP-D only for simplicity.

\begin{figure}[h]
    \centering
    \includegraphics[width=0.4\textwidth]{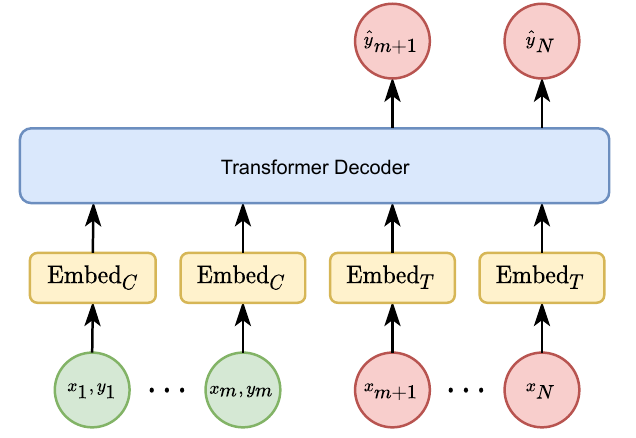}
    \caption{TNPs with two separate embedding layers for the context points and target points.}
    \label{fig:sep_embedding}
\end{figure}

\textbf{Separate embedding layers for context and target points}: The first idea is to use two different embedding layers, one for the context points and one for the target points. The embedding layer for context points will take $(x_{1:C} ,y_{1:C})$ as the input, while the embedding layer for target points will only take $x^*_{1:T}$ as the input. Figure \ref{fig:sep_embedding} depicts this model.

\begin{figure}[h]
    \centering
    \includegraphics[width=0.9\textwidth]{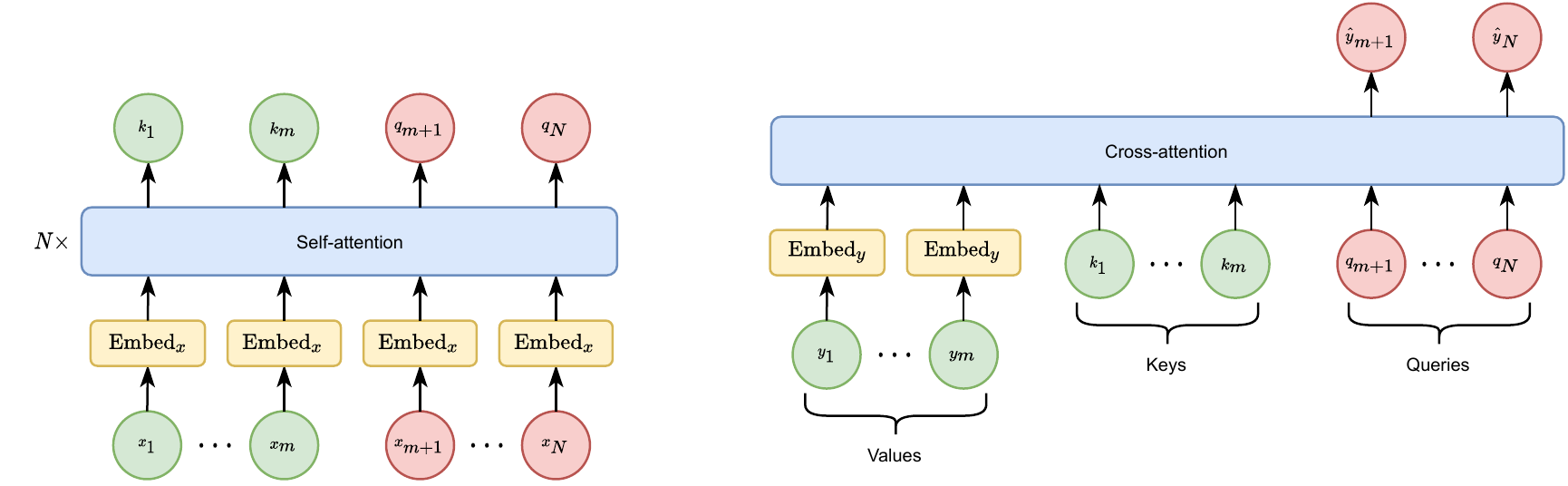}
    \caption{TNPs with cross attention.}
    \label{fig:cross_att}
\end{figure}

\textbf{Cross attention}: Another idea is to use cross-attention, which is depicted in Figure \ref{fig:cross_att}. Specifically, we first implement a stack of self-attention layers to learn the interaction between the $x's$. The outputs of this module will serve as queries and keys in a cross-attention layer, while the context labels $y_{1:C}$ serve as values. This eliminates the need of a mask because the self-attention layer only computes the similarity of $x's$ by using them as queries and keys.

\textbf{Comparison}: Table \ref{tab:alter_input} compares the two alternative input representations with TNP-D. We see that TNP-D is slightly better than Sep-emb, while Cross-att barely works for any evaluation kernel. This is because the model is not allowed to look at pairs of context points $(x_i, y_i)_{i=1}^C$, thus cannot infer the underlying distribution f, or in other words cannot tell which realisation of GP these pairs come from. Therefore, the best thing it can do is to make a random guess with a very large variance. This confirms that coupling $x$ and $y$ to form the input representation is crucial to the performance of TNP.
\begin{table}[h]
\centering
\caption{Comparison of TNP-D vs. the alternative input representations.}
\label{tab:alter_input}
\begin{tabular}{ccccc}
\toprule
Metric                          & Method       & RBF            & Matérn 5/2     & Periodic   \\
\midrule
\multirow{3}{*}{RMSE}           & TNP-D          & $\boldsymbol{0.177 \pm 0.001}$  & $\boldsymbol{0.222 \pm 0.000}$  & $\boldsymbol{0.664 \pm 0.014}$  \\
                                & Sep-emb        & $\boldsymbol{0.177 \pm 0.000}$  & $\boldsymbol{0.222 \pm 0.000}$  & $0.685 \pm 0.006$   \\
                                & Cross-att      & $0.304 \pm 0.122$  & $0.328 \pm 0.112$  & $0.668 \pm 0.050$   \\
\midrule
\multirow{3}{*}{CE}             & TNP-D          & $\boldsymbol{0.043 \pm 0.000}$  & $0.045 \pm 0.000$  & $\boldsymbol{0.129 \pm 0.012}$ \\
                                & Sep-emb        & $0.045 \pm 0.001$  & $\boldsymbol{0.044 \pm 0.001}$  & $0.135 \pm 0.008$   \\
                                & Cross-att      & $0.151 \pm 0.055$  & $0.117 \pm 0.038$  & $0.203 \pm 0.081$   \\
\midrule
\multirow{3}{*}{Log-Likelihood} & TNP-D          & $\boldsymbol{1.388 \pm 0.004}$  & $\boldsymbol{0.954 \pm 0.005}$  & $-3.525 \pm 0.367$    \\
                                & Sep-emb        & $1.375 \pm 0.004$  & $\boldsymbol{0.954 \pm 0.003}$  & $\boldsymbol{-3.232 \pm 0.238}$   \\
                                & Cross-att      & $0.115 \pm 0.337$  & $-0.012 \pm 0.281$  & $-2.989 \pm 1.225$   \\
\bottomrule
\end{tabular}
\end{table}

\subsection{Autoregressive Transformer Neural Processes}
There are other architectures that enable autoregressiveness in addition to what was presented in Section \ref{sec:tnp_a}. We present those architectures here, their differences with TNP-A, and their relative performances.
\begin{figure}[ht]
     \centering
     \begin{subfigure}[b]{0.47\textwidth}
         \centering
         \includegraphics[width=\textwidth]{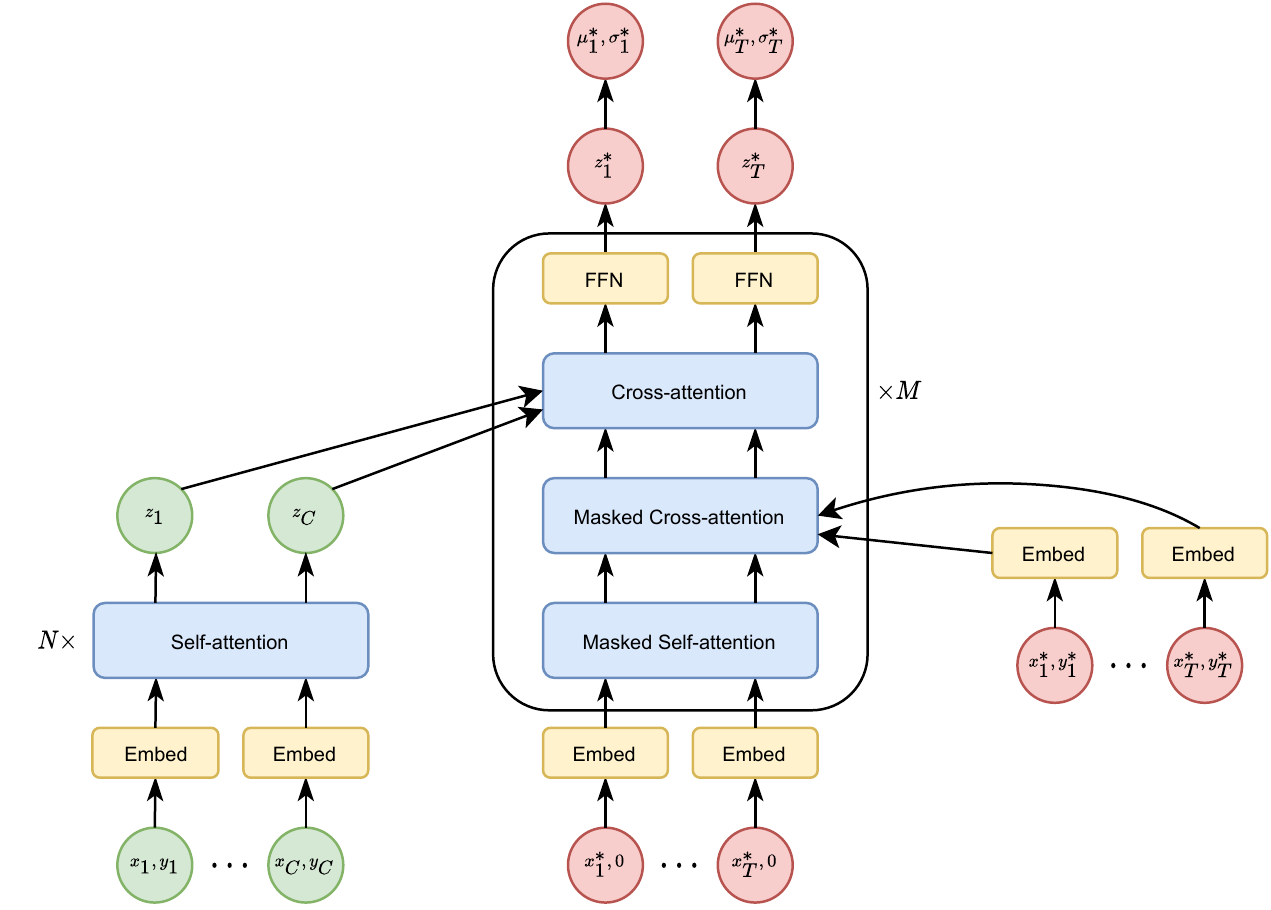}
         \caption{TNP-A-1}
         \label{fig:tnp_a_alter_1}
     \end{subfigure}
     \begin{subfigure}[b]{0.47\textwidth}
         \centering
         \includegraphics[width=\textwidth]{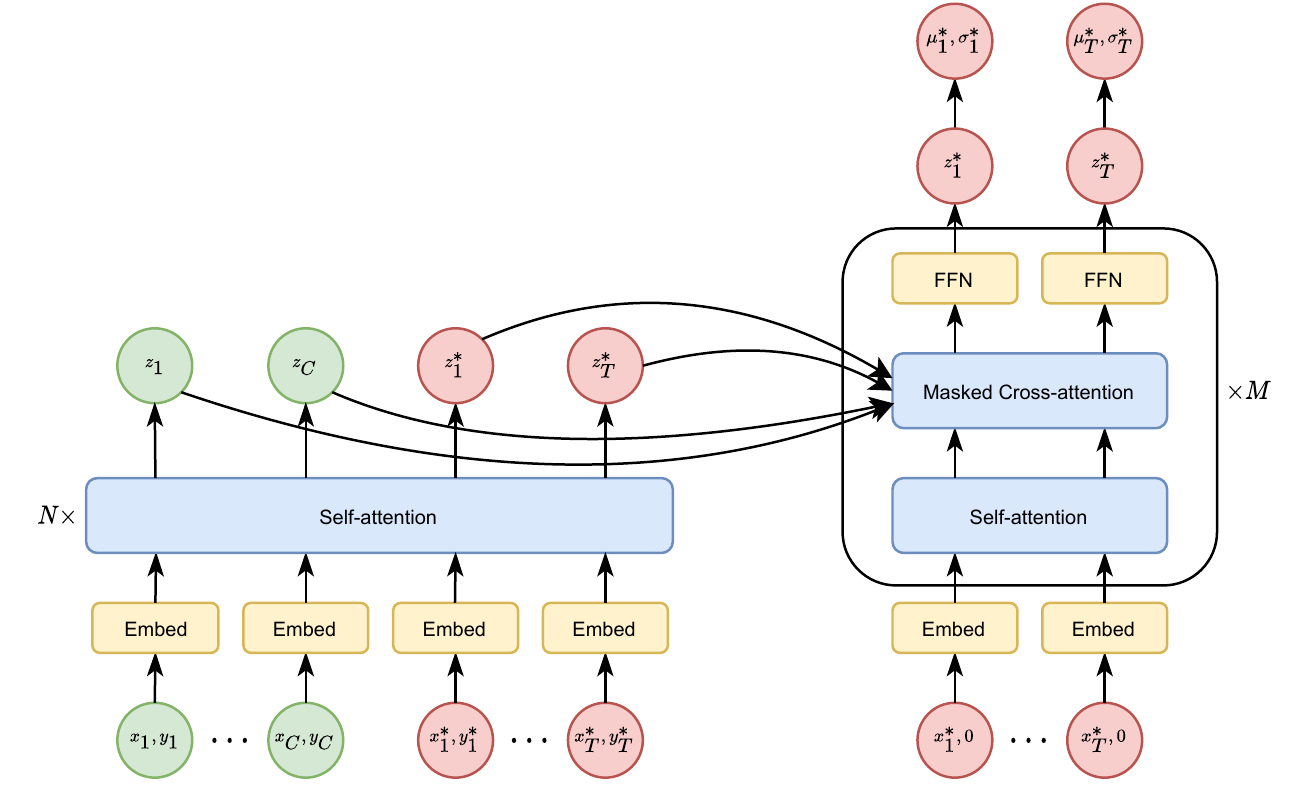}
         \caption{TNP-A-2}
         \label{fig:tnp_a_alter_2}
     \end{subfigure}
    \caption{The alternative architectures for TNP-A}
    \label{fig:tnp_a_alter}
\end{figure}

Figure \ref{fig:tnp_a_alter} shows the two alternative architectures. In TNP-A-1 (Figure \ref{fig:tnp_a_alter_1}), we encode the context points and the target points separately, and then use a stack of masked self-attention, masked-cross attention, and cross-attention layers that allow each target point to attend all the context points and the previous target points. In each masked layer, we ensure that the component at position $i^{th}$ cannot attend positions $j^{th}, j > i$. Note that in this model, there is no interaction between the context points and the target points.

In the TNP-A-2 (Figure \ref{fig:tnp_a_alter_2}), we allow this interaction by feeding all context points and target points into the same encoder. In the encoder, we use a masking mechanism such that the context points do not attend the target points, as in evaluation we do not have ground-truth but only predicted target. Similarly to TNP-A-2, the masked cross-attention module in Pure GPT 2 stops the component at position $i^{th}$ from attending positions $j^{th}$ with $j > i$.

\textbf{Comparison}: We compare the relative performances of TNP, TNP-A-1, and TNP-A-2 in the meta regression task. Table \ref{tab:alter_autoregress} shows the results.
\begin{table}[h]
\centering
\caption{Comparison of TNP-A vs. the alternative autoregressive models.}
\label{tab:alter_autoregress}
\begin{tabular}{ccccc}
\toprule
Metric                          & Method       & RBF            & Matérn 5/2     & Periodic   \\
\midrule
\multirow{3}{*}{RMSE}           & TNP-A          & $\boldsymbol{0.178 \pm 0.000}$  & $\boldsymbol{0.222 \pm 0.000}$  & $\boldsymbol{0.660 \pm 0.002}$   \\
                                & TNP-A-1        & $0.182 \pm 0.001$  & $0.225 \pm 0.000$  & $0.673 \pm 0.007$   \\
                                & TNP-A-2        & $0.184 \pm 0.002$  & $0.223 \pm 0.001$  & $0.667 \pm 0.008$   \\
\midrule
\multirow{3}{*}{CE}             & TNP-A          & $0.045 \pm 0.000$  & $0.044 \pm 0.000$  & $\boldsymbol{0.119 \pm 0.008}$  \\
                                & TNP-A-1        & $\boldsymbol{0.044 \pm 0.000}$  & $0.044 \pm 0.000$  & $0.147 \pm 0.009$   \\
                                & TNP-A-2        & $0.046 \pm 0.002$  & $\boldsymbol{0.042 \pm 0.001}$  & $0.168 \pm 0.009$   \\
\midrule
\multirow{3}{*}{Log-Likelihood} & TNP-A          & $\boldsymbol{1.628 \pm 0.001}$  & $\boldsymbol{1.207 \pm 0.003}$  & $\boldsymbol{-2.257 \pm 0.168}$  \\
                                & TNP-A-1        & $1.477 \pm 0.016$  & $1.070 \pm 0.018$  & $-3.380 \pm 0.314$   \\
                                & TNP-A-2        & $1.577 \pm 0.024$  & $1.167 \pm 0.018$  & $-4.607 \pm 0.390$   \\
\bottomrule
\end{tabular}
\end{table}

\subsection{Low-rank approximation for TNP-ND}
\begin{figure}[h!]
    \centering
    \includegraphics[width=0.32\textwidth]{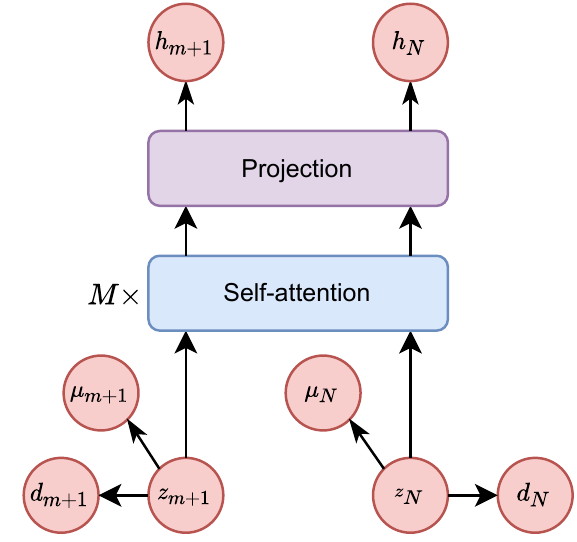}
    \caption{The decoder architecture of the low-rank version of TNP-ND.}
    \label{fig:TNP-ND_lowrank}
\end{figure}
As mentioned in Section \ref{sec:tnp_nd}, we can use Low-rank approximation to parameterize the covariance matrix in the predictive distribution $\mathcal{N}(y^*_{1:T} \mid \mu_\theta(x^*_{1:T}, C), \Sigma_\theta(x^*_{1:T}, C))$. Figure \ref{fig:TNP-ND_lowrank} depicts the parameterization, which is similar to TNP-ND, except that we have an additional MLP that outputs $d_{m+1:N}$. The covariance is computed as:
\begin{equation}
    \Sigma = H H^T + \exp(D), \ H \in \mathbb{R}^{n \times p}, \ D \in \mathbb{R}^{n \times n},
\end{equation}
in which $D$ is a diagonal matrix formed by $d_{m+1:N}$. This parameterization guarantees that $\Sigma$ is a positive definite matrix.

\paragraph{Results} We compare TNP-ND with Cholesky decomposition vs TNP-ND with low-rank approximation in the 1-D regression problem. Table \ref{tab:1d_lowrank} shows the results.
\begin{table}[h!]
\centering
\caption{Comparison of TNP-ND with Cholesky decomposition vs TNP-ND with low-rank approximation on various GP kernels and evaluation metrics. We train $5$ instances with different seeds for each model and report the mean and std.}
\label{tab:1d_lowrank}
\scalebox{0.85}{
\begin{tabular}{ccccc}
\toprule
Metric                          & Method       & RBF            & Matérn 5/2     & Periodic   \\
\midrule
\multirow{2}{*}{RMSE}           & TNP-ND-Cholesky & $\boldsymbol{0.180 \pm 0.001}$  & $\boldsymbol{0.223 \pm 0.000}$  & $\boldsymbol{0.670 \pm 0.009}$  \\
                                & TNP-ND-Lowrank  & $0.186 \pm 0.001$  & $0.225 \pm 0.001$  & $0.680 \pm 0.010$ \\
\midrule
\multirow{2}{*}{CE}             & TNP-ND-Cholesky & $0.048 \pm 0.001$  & $0.050 \pm 0.001$  & $0.155 \pm 0.009$  \\
                                & TNP-ND-Lowrank  & $\boldsymbol{0.043 \pm 0.001}$  & $\boldsymbol{0.044 \pm 0.001}$  & $\boldsymbol{0.153 \pm 0.019}$  \\
\midrule
\multirow{2}{*}{Log-Likelihood} & TNP-ND-Cholesky & $1.46 \pm 0.00$  & $1.02 \pm 0.00$  & $\boldsymbol{-4.13 \pm 0.33}$   \\
                                & TNP-ND-Lowrank & $\boldsymbol{1.565 \pm 0.007}$  & $\boldsymbol{1.117 \pm 0.003}$  & $-5.702 \pm 0.509$  \\
\bottomrule
\end{tabular}
}
\end{table}

\section{Pretraining TNPs}
In the main results of the paper, we trained TNPs in a meta-training regime, where the model makes predictions for the target points conditioning on the context. This was explicitly designed to match the evaluation setting. However, previous studies~\cite{brown2020language,chan2022data} have shown that the capacity for in-context learning in transformer-based models is emergent. That is, the model is capable of performing few-shot learning, without being explicitly trained to do so. We conduct experiments to investigate this emergent in-context learning behavior of TNPs. Specifically, during training, the model observes sequences of evaluations $(x_{1:N}, y_{1:N})$, and learns to predict each point given the preceding points:
\begin{align}
 \mathcal{L}(\theta) = \mathbb{E}_{x_{1:N}, y_{1:N}}[\log p_\theta(y_{1:N}|x_{1:N})] =  \mathbb{E}_{x_{1:N}, y_{1:N}}\left[\sum_{m=1}^N \log p_\theta(y_m | x_{1:m}, y_{1:m-1})\right]
\end{align}
We term the variant trained using this autoregressive objective TNP-A-Pretrained. During evaluation, we introduce the notions of context and target via proper masking, where each target conditions on the context points and the preceding predicted target points. This resembles TNP-A, and the only difference is in the training phase. We compare TNP-A-Pretrained to TNP-A, TNP-D, and TNP-ND in all $4$ meta-learning tasks below.

\newpage
\subsection{1D Regression}
\begin{table}[ht]
\centering
\caption{Comparison of TNP-A-Pretrained vs three TNP variants with various evaluation metrics. We train $5$ instances with different seeds for each model and report the mean and std.}
\label{tab:regression_pretrained}
\scalebox{1.0}{
\begin{tabular}{ccccc}
\toprule
Metric                          & Method       & RBF            & Matérn 5/2     & Periodic   \\
\midrule
\multirow{4}{*}{RMSE}           & TNP-D         & $\boldsymbol{0.177 \pm 0.001}$  & $\boldsymbol{0.222 \pm 0.000}$  & $0.664 \pm 0.014$  \\
                                & TNP-A          & $0.178 \pm 0.000$  & $\boldsymbol{0.222 \pm 0.000}$  & $\boldsymbol{0.660 \pm 0.002}$   \\
                                & TNP-ND & $0.180 \pm 0.001$  & $0.223 \pm 0.000$  & $0.670 \pm 0.009$  \\
                                & TNP-A-Pretrained & $0.180 \pm 0.001$ & $0.224 \pm 0.001$ & $0.679 \pm 0.007$ \\
\midrule
\multirow{4}{*}{CE}             & TNP-D         & $\boldsymbol{0.043 \pm 0.000}$  & $0.045 \pm 0.000$  & $0.129 \pm 0.012$ \\
                                & TNP-A          & $0.045 \pm 0.000$  & $\boldsymbol{0.044 \pm 0.000}$  & $\boldsymbol{0.119 \pm 0.008}$  \\
                                & TNP-ND & $0.048 \pm 0.001$  & $0.050 \pm 0.001$  & $0.155 \pm 0.009$  \\
                                & TNP-A-Pretrained & $\boldsymbol{0.043 \pm 0.005}$ & $0.052 \pm 0.007$ & $0.148 \pm 0.020$ \\
\midrule
\multirow{4}{*}{Log-Likelihood} & TNP-D         & $1.39 \pm 0.00$  & $0.95 \pm 0.01$  & $-3.53 \pm 0.37$    \\
                                & TNP-A          & $\boldsymbol{1.63 \pm 0.00}$  & $\boldsymbol{1.21 \pm 0.00}$  & $\boldsymbol{-2.26 \pm 0.17}$  \\
                                & TNP-ND & $1.46 \pm 0.00$  & $1.02 \pm 0.00$  & $-4.13 \pm 0.33$   \\
                                & TNP-A-Pretrained & $\boldsymbol{1.63 \pm 0.01}$ & $1.19 \pm 0.03$ & $-3.69 \pm 0.92$ \\
\bottomrule
\end{tabular}
}
\end{table}

\subsection{Image Completion}
\begin{table}[h!]
\centering
\caption{Comparison of TNP-A-Pretrained vs three TNP variants on CelebA dataset with various evaluation metrics. We train $5$ instances with different seeds for each model and report the mean and std.}
\label{tab:celeba_pretrained}
\scalebox{1.0}{
\begin{tabular}{cccc}
\toprule
Method       & RMSE          & Log-likelihood \\
\midrule
TNP-D        & $0.112 \pm 0.000$ & $3.891 \pm 0.006$ \\ 
TNP-A        & $0.115 \pm 0.000$ & $\boldsymbol{5.818 \pm 0.011}$  \\
TNP-ND       & $\boldsymbol{0.111 \pm 0.000}$ & $5.477 \pm 0.016$  \\
TNP-A-Pretrained & $0.114 \pm 0.000$ & $4.457 \pm 0.011$ \\
\bottomrule
\end{tabular}
}
\end{table}

\begin{table}[h!]
\centering
\caption{Comparison of TNP-A-Pretrained vs three TNP variants on EMNIST dataset with various evaluation metrics. We train $5$ instances with different seeds for each model and report the mean and std. We evaluate on both seen and unseen classes.}
\label{tab:emnist_pretrained}
\begin{tabular}{@{}cccc@{}}
\toprule
Setting                                                                           & Method       & RMSE           & Log-likelihood \\
\midrule
\multirow{4}{*}{\begin{tabular}[c]{@{}c@{}}Seen classes\\ (0-9)\end{tabular}}     & TNP-D & $0.119 \pm 0.002$  & $1.461 \pm 0.010$  \\
                                                                                  & TNP-A  & $0.122 \pm 0.001$  & $\boldsymbol{1.537 \pm 0.005}$  \\
                                                                                  & TNP-ND & $\boldsymbol{0.116 \pm 0.000}$ & $1.497 \pm 0.002$  \\
                                                                                  & TNP-A-Pretrained & $0.130 \pm 0.000$ & $1.497 \pm 0.002$ \\
\midrule
\multirow{4}{*}{\begin{tabular}[c]{@{}c@{}}Unseen classes\\ (10-46)\end{tabular}} & TNP-D & $\boldsymbol{0.139 \pm 0.001}$  & $1.308 \pm 0.003$ \\
                                                                                  & TNP-A & $0.142 \pm 0.001$  & $\boldsymbol{1.413 \pm 0.005}$  \\
                                                                                  & TNP-ND & $0.140 \pm 0.001$ & $1.314 \pm 0.004$  \\
                                                                                  & TNP-A-Pretrained & $0.159 \pm 0.000$ & $1.256 \pm 0.008$ \\
\bottomrule
\end{tabular}
\end{table}

\newpage
\subsection{Contextual Bandits}
\begin{table*}[ht]
\centering
\caption{Comparison of TNP-A-Pretrained vs three TNP variants on cumulative regret on contextual bandit problems with different values of $\delta$. We run each model $50$ times for each value of $\delta$ and report the mean and std.}
\label{tab:contextual_bandits_pretrained}
\scalebox{0.85}{
\begin{tabular}{cccccccc}
\toprule
Method       & $\delta=0.7$   & $\delta=0.9$   & $\delta=0.95$  & $\delta=0.99$  & $\delta=0.995$  & $\delta=0.999$  & Average  \\ 
\midrule
Uniform      & $100.00 \pm 1.18$  & $100.00 \pm 3.03$  & $100.00 \pm 4.16$  & $100.00 \pm 7.52$  & $100.00 \pm 8.11$   & $100.00 \pm 7.96$  & $100.00 \pm 5.97$   \\ 
TNP-D & $\boldsymbol{1.18 \pm 0.94}$    & $1.70 \pm 0.41$    & $2.55 \pm 0.43$    & $\boldsymbol{3.57 \pm 1.22}$    & $\boldsymbol{4.68 \pm 1.09}$     & $9.56 \pm 0.44$    & $\boldsymbol{3.87 \pm 2.91}$     \\
TNP-A          & $3.67 \pm 4.88$ & $4.04 \pm 2.38$ & $4.29 \pm 2.36$ & $5.79 \pm 5.27$ & $9.29 \pm 7.62$ & $\boldsymbol{6.13 \pm 2.50}$ & $5.54 \pm 4.98$ \\
TNP-ND & $1.76 \pm 0.61$ & $\boldsymbol{1.41 \pm 0.98}$ & $\boldsymbol{1.61 \pm 1.65}$ & $4.98 \pm 2.84$ & $7.22 \pm 3.28$ & $13.66 \pm 2.92$ & $5.11 \pm 4.94$ \\
TNP-A-Pretrained & $1.53 \pm0.09$ & $4.96 \pm0.17$ & $8.00 \pm0.13$ & $23.98 \pm0.42$ & $34.35 \pm0.01$ & $83.79 \pm0.02$ & $26.10 \pm 28.22$ \\
\bottomrule
\end{tabular}
}
\end{table*}

\subsection{Bayesian Optimization}
\begin{figure*}[h]
    \centering
    \includegraphics[width=0.7\textwidth]{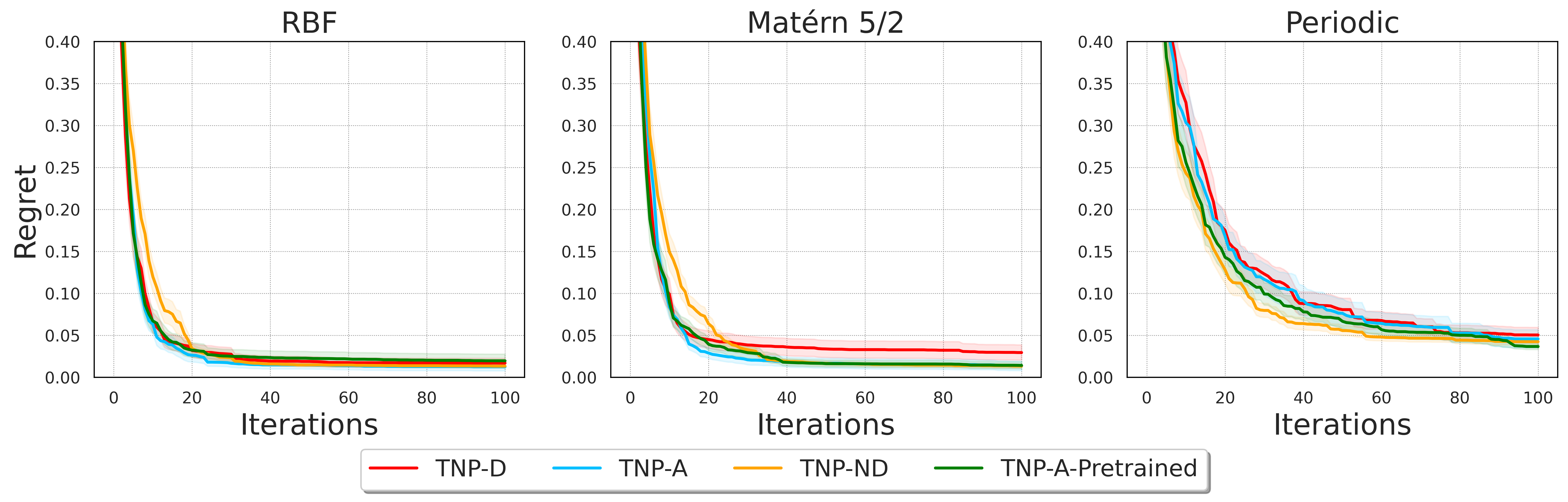}
    \caption{Regret performance on 1D BO tasks of TNP-A-Pretrained and the three TNP variants. For each kernel, we generate $100$ functions and report the mean and standard deviation.}
    \label{fig:bo_1d_pretrained}
\end{figure*} 
\begin{figure*}[h!]
    \centering
    \includegraphics[width=0.7\textwidth]{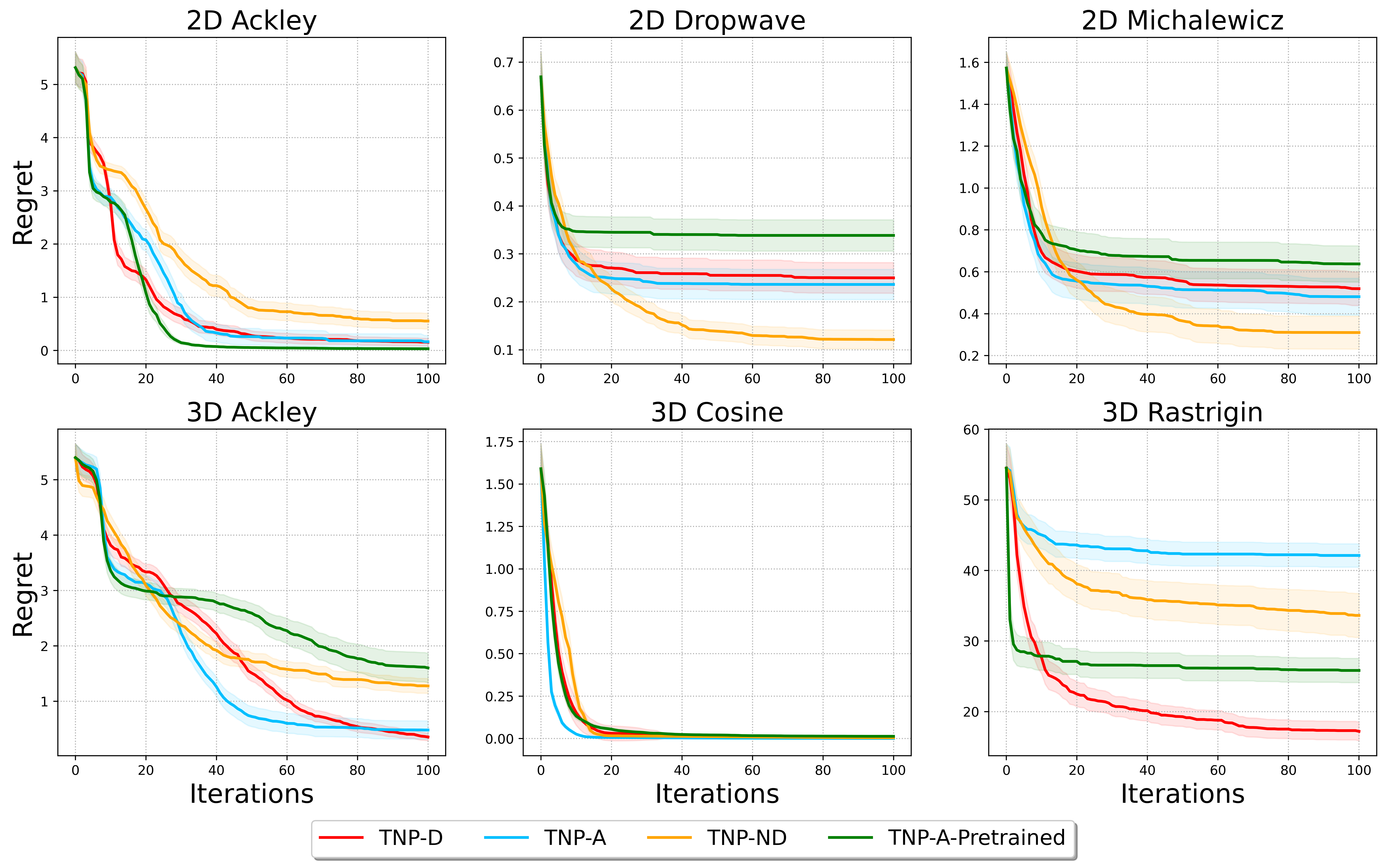}
    \caption{Regret performance on 2D and 3D BO tasks of TNP-A-Pretrained and the three TNP variants. For each function, we run BO for $100$ times with different seeds and report the mean and standard deviation.}
    \label{fig:bo_multi_pretrained}
\end{figure*}

Overall, TNP-A-Pretrained performs reasonably well in all tasks, but still lags behind the meta-trained TNPs, especially in the two sequential decision making tasks. Improving this pretraining scheme to match the performance of the meta-trained models is an interesting research direction for future work.

\end{document}